\documentclass[12pt]{amsart}

\PassOptionsToPackage{numbers, compress}{natbib}

\usepackage{amssymb,amscd,amsthm, verbatim,amsmath,color,fancyhdr, mathrsfs}
\usepackage{graphicx}
\usepackage{turnstile}
\usepackage[utf8]{inputenc} % allow utf-8 input
\usepackage[T1]{fontenc}    % use 8-bit T1 fonts
\usepackage{hyperref}       % hyperlinks
\usepackage{url}            % simple URL typesetting
\usepackage{booktabs}       % professional-quality tables
\usepackage{amsfonts}       % blackboard math symbols
\usepackage{nicefrac}       % compact symbols for 1/2, etc.
\usepackage{microtype}      % microtypography
\usepackage{xcolor}         % colors

\usepackage{algorithm}
\usepackage{algpseudocode}
\usepackage{subcaption}
\usepackage{mathtools}
\usepackage{amsthm}
\usepackage{graphicx}
\usepackage{wrapfig}
\usepackage{enumitem}

\usepackage[letterpaper, left=2.5cm, right=2.5cm, top=2.5cm,
bottom=2.5cm,dvips]{geometry}

\newtheorem{theorem}{Theorem}[section]

\newtheorem{definition}[theorem]{Definition}

\title{Atlas flow : compatible local structures on the manifold}

\author{Taejin Paik}
\author{Jaemin Park}
\author{Jung Ho Park}

\begin{document}
\maketitle

\vskip.5in

\begin{abstract}
In this paper, we focus on the intersections of a manifold's local structures to analyze the global structure of a manifold. 
We obtain local regions on data manifolds such as the latent space of StyleGAN2, using Mapper, a tool from topological data analysis. 
We impose gluing compatibility conditions on overlapping local regions, which guarantee that the local structures can be glued together to the global structure of a manifold. 
We propose a novel generative flow model called \textit{Atlas flow} that uses compatibility to reattach the local regions. 
Our model shows that the generating processes perform well on synthetic dataset samples of well-known manifolds with noise. 
Furthermore, we investigate the style vector manifold of StyleGAN2 using our model.
\end{abstract}

%%%%%%%%%%%%%%%%%%%%%%%%%%%%%%%%%%%%%%%%%%%%%%%%%%%%%%%%%%%%%%%%%%%%%%%%%%%%%%%%%%%%%%%%%%%

\section{Introduction}
A popular method to improve data-driven models such as deep neural networks improves the data representation using manifold learning and density estimation.

Various methods have been proposed for manifold learning assuming the manifold hypothesis \cite{fefferman2016testing}. 
In statistics, dimension reduction methods, such as PCA, Isomap \cite{tenenbaum2000global}, and tSNE \cite{van2008visualizing}, have been widely used to infer the latent manifold.
Furthermore, various deep neural networks have been studied to apply manifold learning techniques such as autoencoder \cite{baldi2012autoencoders, kingma2014auto, kingma2016improved, malhotra2016lstm, masci2011stacked}.

On the other hand, density estimation is as important as manifold learning. 

Generative models are practical examples of methods that learn density either implicitly \cite{goodfellow2014generative, karras2019style, karras2020analyzing, karras2021alias, kingma2014auto}, or explicitly \cite{rezende2015variational, kingma2018glow}.

Recently, a flow-based framework\cite{brehmer2020flows} was proposed, called manifold-learning flow to perform both manifold learning and density estimation.
In this setting, there are two flow-based maps: one for manifold learning, and one for density estimation.
Using these two maps, one can often identify the full data manifold and generate sample points on this manifold using the density function.
To improve this philosophy, we studied a new flow-based model for a general manifold which includes spaces such as spheres.

In order to analyze a manifold with a complex shape, theoretically, we have to describe the manifold with the \textit{atlas} \cite{lee2013smooth}.
An atlas refers to a collection of coordinate charts, which are local structures consisting of local regions and maps. 
A coordinate map is a diffeomorphism from a local region of a manifold to a Euclidean space preserving topological and geometrical properties.
Using this approach, several studies \cite{cohn2021topologically, kalatzis2021multi} trained models to find the local parts of a manifold.
However, these studies used partitions as multi-charts, so their components trained on local regions are independent of each other.
In short, these studies investigated different non-overlapping components yielding independent local structures. 

Hence, we study our method to match the concept of the manifold with coordinate charts that have overlapping local regions appropriately.
By allowing overlap of regions, trained local structures can be combined and hence used to cover the entire data manifold.
Thus, it should be noted that allowing overlapping regions is a more general approach than partitioning (even if they both consider local geometry).
We adopt an idea of topological data analysis (TDA), namely Mapper, in order to select appropriate local regions.
When a prior data manifold is unknown, Mapper is used to infer the global topological structure of a manifold by considering the connectivity of local regions.

%%%%%%%%%%%%%%%%%%%%%%
We suggest the term ``compatibility" in order to describe the smoothness of overlaps of mutually trained and distinct local structures.
Through a training process for coordinate maps, we intend reconstruction points to coincide for compatible regions.
Moreover, we also study how to decompose densities on overlapping regions using disintegration, which is a notion of probability theory.
In section \ref{sec:method}, we explain our story of study and propose a novel method \textit{Atlas flow}.
In section \ref{sec:exp}, we conduct two types of experiments.

We first apply our method to synthetic datasets which consist of sample points on manifolds embedded in $\mathbb{R}^3$, and then compare to the original datasets. % 수정
Moreover, we apply our method to the latent space of StyleGAN2 which was pre-trained on FFHQ dataset. Our contributions are as follows:

\begin{enumerate}[leftmargin=*]
    \item Our method is a generative model working under the assumption that the dataset is lying on a manifold.
    In principle, we can handle even complicated manifolds using our Atlas flow model.
    \item This includes a new manifold learning algorithm that works for a wider class of manifolds. 
    Compatibility ensures successful inference of the overall manifold by combining local regions. 
    \item Furthermore, we also obtain a measure-theoretic decomposition method of a density function using the concept of disintegration. 
    Each decomposed density is supported on a local region obtained from manifold learning.
    Compatibility ensures successful inference of the overall density function by integrating decomposed densities. 
\end{enumerate}

\section {Related works}

\subsection{Flow-based models}
% General explanation
Normalizing flow (NF) \cite{rezende2015variational, gemici2016normalizing, kingma2018glow, durkan2019neural} is one of the most elegant frameworks in generative models. 
NF can be considered as an approach to transforming a trivial density into a non-trivial one. 
The essence of NF is that its encoder is a diffeomorphism. 
A diffeomorphism means a smooth bijection whose inverse is also smooth.
By the definition of diffeomorphism, the decoder of NF becomes the inverse map of the encoder.
Thus NF can preserve geometrical information between a data manifold and corresponding latent space. 
Moreover, diffeomorphisms have the nice property that their compositions are also diffeomorphisms. 
This property enhances the ability of NF to learn non-trivial density.

Since the first NF assumes the data manifold is a Euclidean space, 
a branch of study \cite{gemici2016normalizing, rezende2020normalizing, mathieu2020riemannian} generalizes the latent manifold from a Euclidean space to Riemannian manifolds. 
Thus, one can estimate the appropriate density on a manifold when prior knowledge of the data manifold exists.
The manifold-learning flow \cite{brehmer2020flows} is based on these ideas and combines manifold learning and density learning. 
Manifold-learning flow consists of two parts: outer and inner networks. 
The outer network is an autoencoder based on NF and learns the manifold structure. 
The inner network learns the density which is tractable in this setting. 

\subsection{Multi-Chart Manifold Learning} 

We need to consider a collection of local domains covering the manifold; we allow overlap in this covering.
Hence the knowledge of such a collection is of significant interest to many researchers.
Estimating the latent dimension has been actively studied by many researchers.
Nowadays, some research \cite{cohn2021topologically, kalatzis2021multi} suggests that studying a collection of local parts is worthwhile. 
Topologically-informed atlas learning \cite{cohn2021topologically} splits a manifold to preserve topological information such as $k$-dimensional holes. 
Multi-chart flows \cite{kalatzis2021multi} which is similar to our work divides a manifold into local partitions and learns each coordinate map using the Manifold-learning flow \cite{brehmer2020flows}. 
We propose an algorithm to analyze a manifold by using a collection of local domains, not partitions. 
This makes our work different from Multi-chart flows and necessary to deal with overlapping charts. 

\begin{figure*}[t]
\vskip 0.2in
\begin{center}
\centerline{\includegraphics[width=0.9\textwidth]{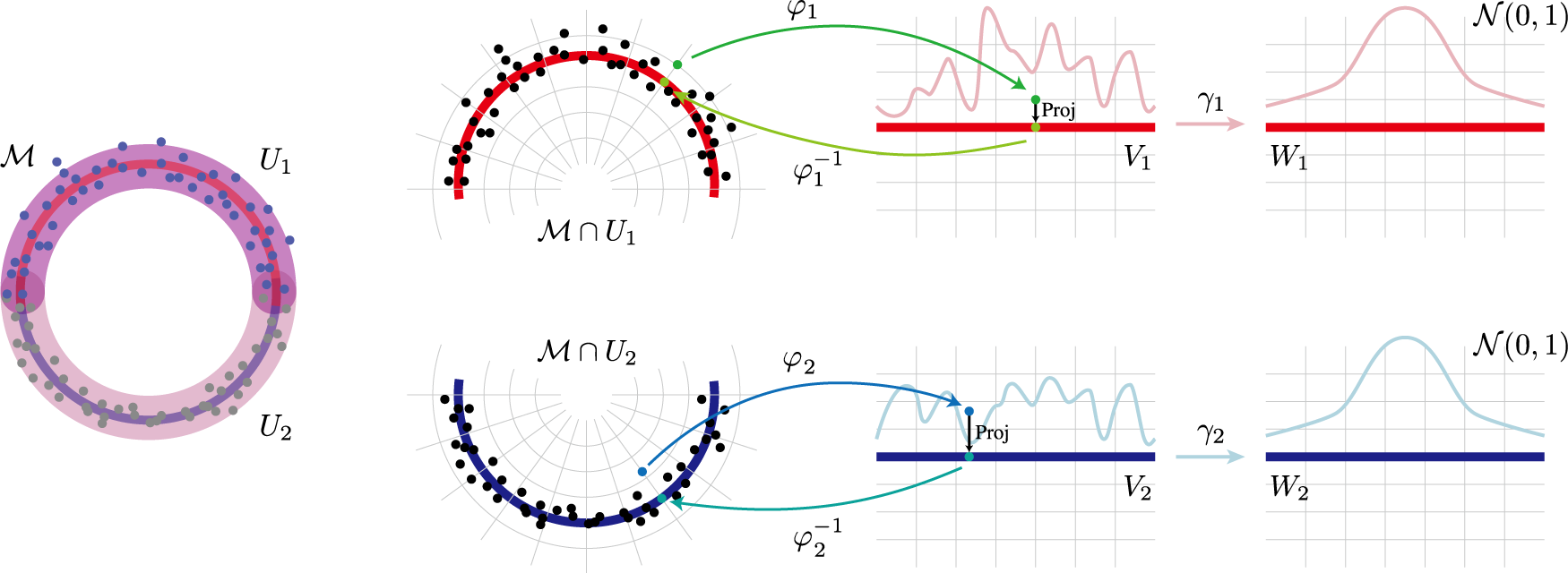}}
\caption{Framework scheme of Atlas flow. A data manifold is covered by coordinate domains obtained from Mapper. Each coordinate domain $\mathcal{M} \cap U_k$ maps to $V_k \subset \mathbb{R}^n$ and reconstructed points are exactly on the submanifold. Finally, $V_k$ maps to $W_k$ with tractable density (standard normal). Generating process is in reverse order: sampling points from $W_k$ and mapping to $\mathcal{M}$ by $\varphi_k^{-1} \circ \gamma_k^{-1}$.}
\label{method}
\end{center}
\vskip -0.2in
\end{figure*}

\section{Method}\label{sec:method}

\subsection{Motivation}

\begin{figure}[ht]
\vskip 0in
\begin{center}
\begin{subfigure}[b]{0.32\columnwidth}
    \centering
    \centerline{\includegraphics[width=\columnwidth]{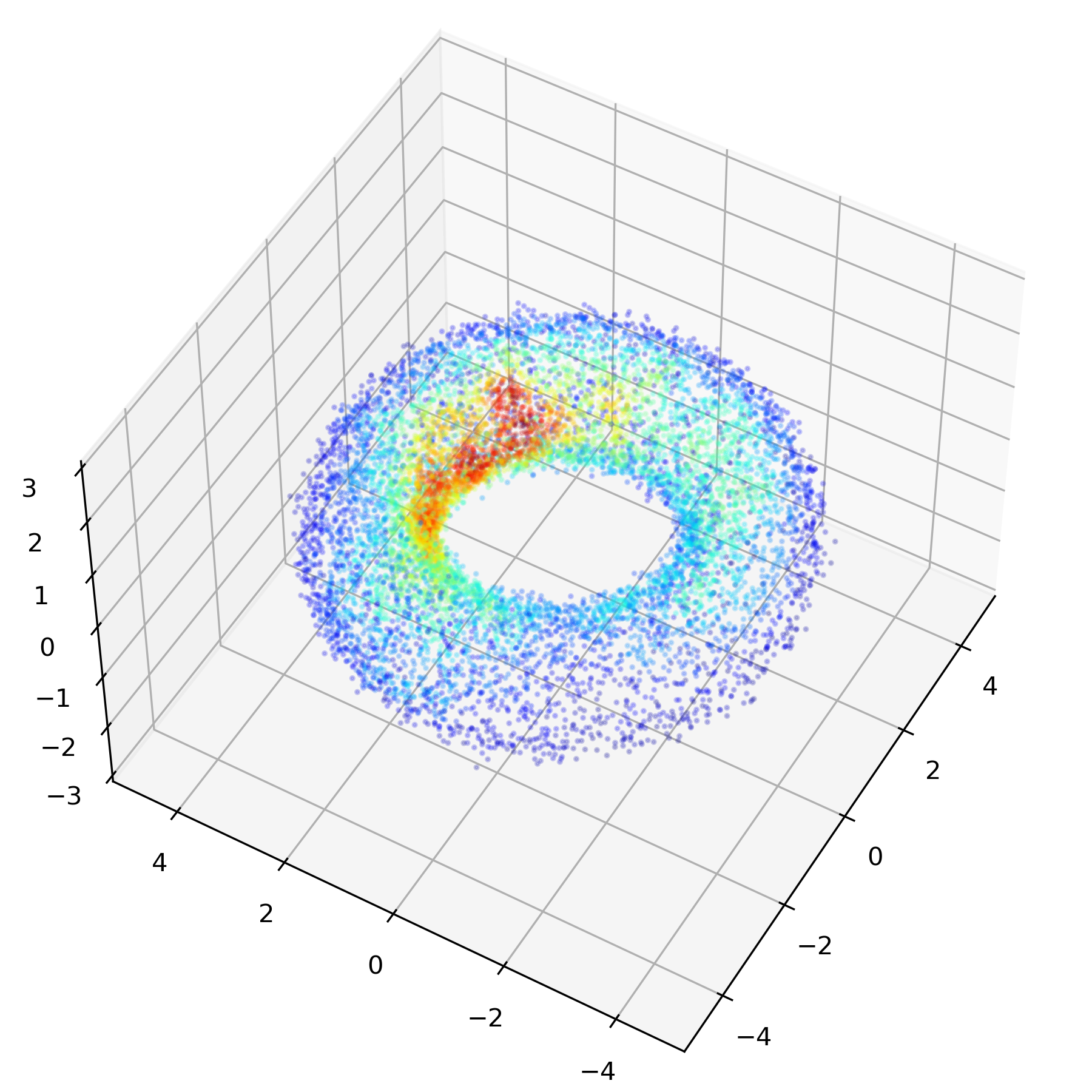}}
    \caption{}
\end{subfigure}
\begin{subfigure}[b]{0.32\columnwidth}
    \centering
    \centerline{\includegraphics[width=\columnwidth]{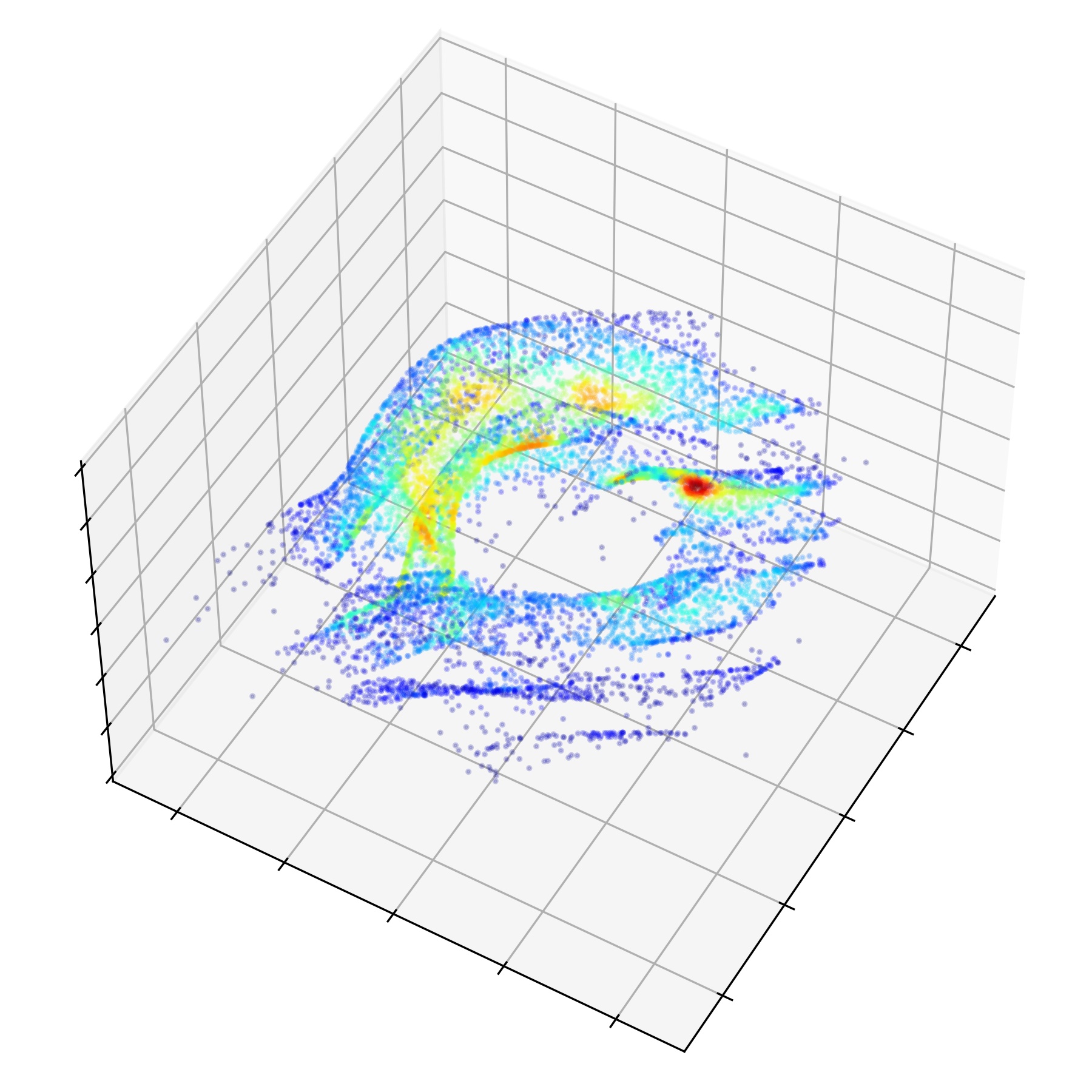}}
    \caption{}
\end{subfigure}
\begin{subfigure}[b]{0.32\columnwidth}
    \centering
    \centerline{\includegraphics[width=\columnwidth]{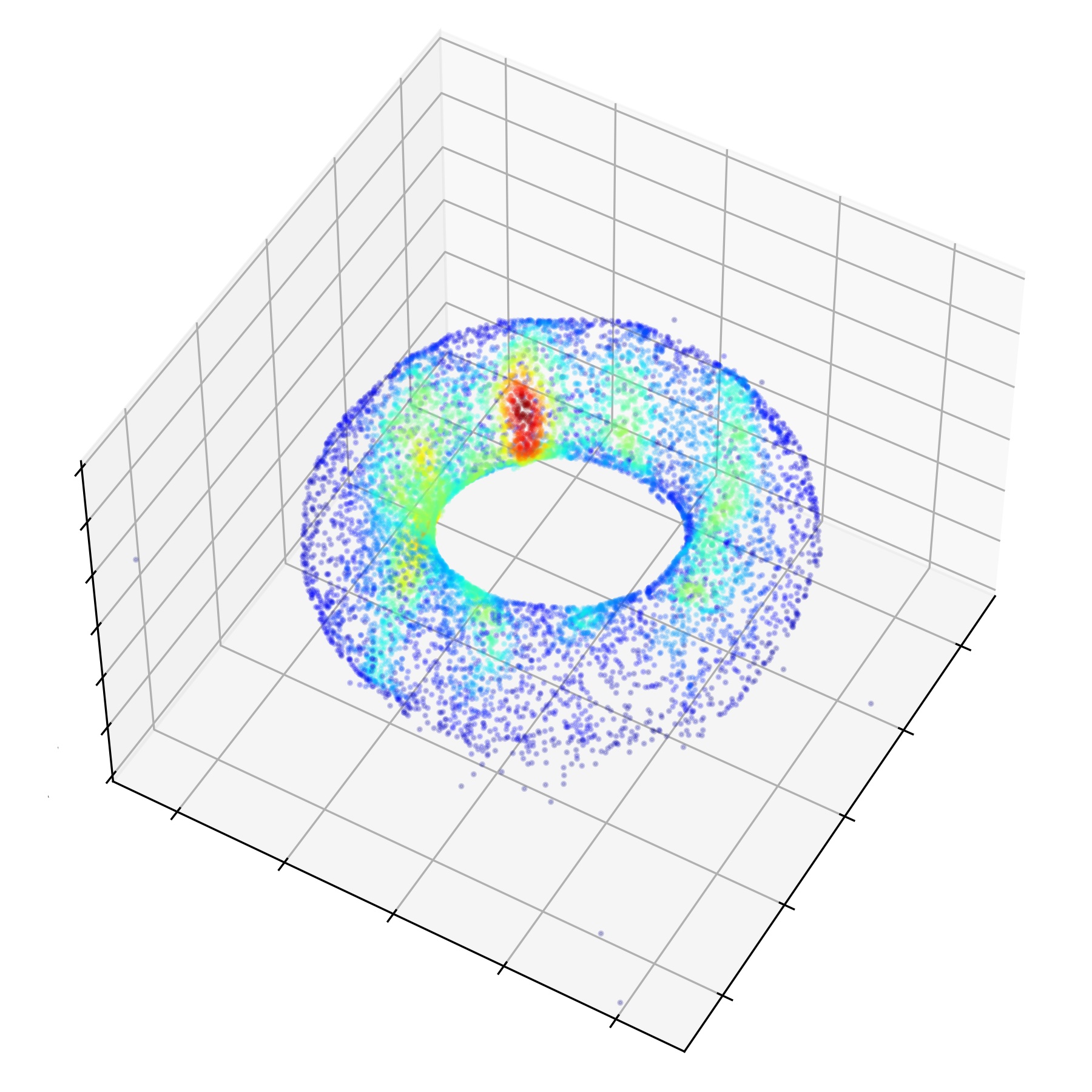}}
    \caption{}
\end{subfigure}

\begin{subfigure}[b]{0.35\columnwidth}
    \centering
    \centerline{\includegraphics[width=\columnwidth]{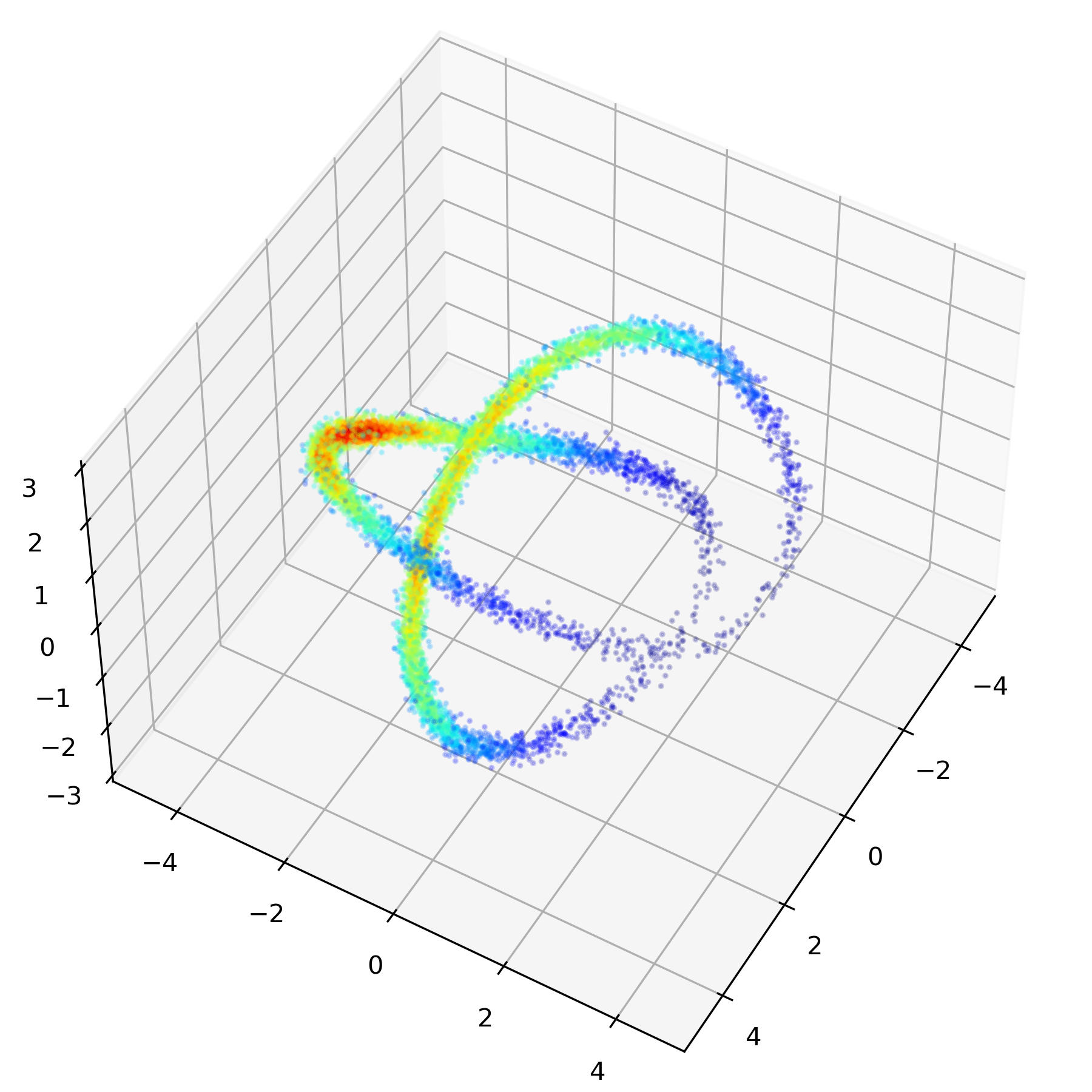}}        
    \caption{}
\end{subfigure}
\begin{subfigure}[b]{0.35\columnwidth}
    \centering
    \centerline{\includegraphics[width=\columnwidth]{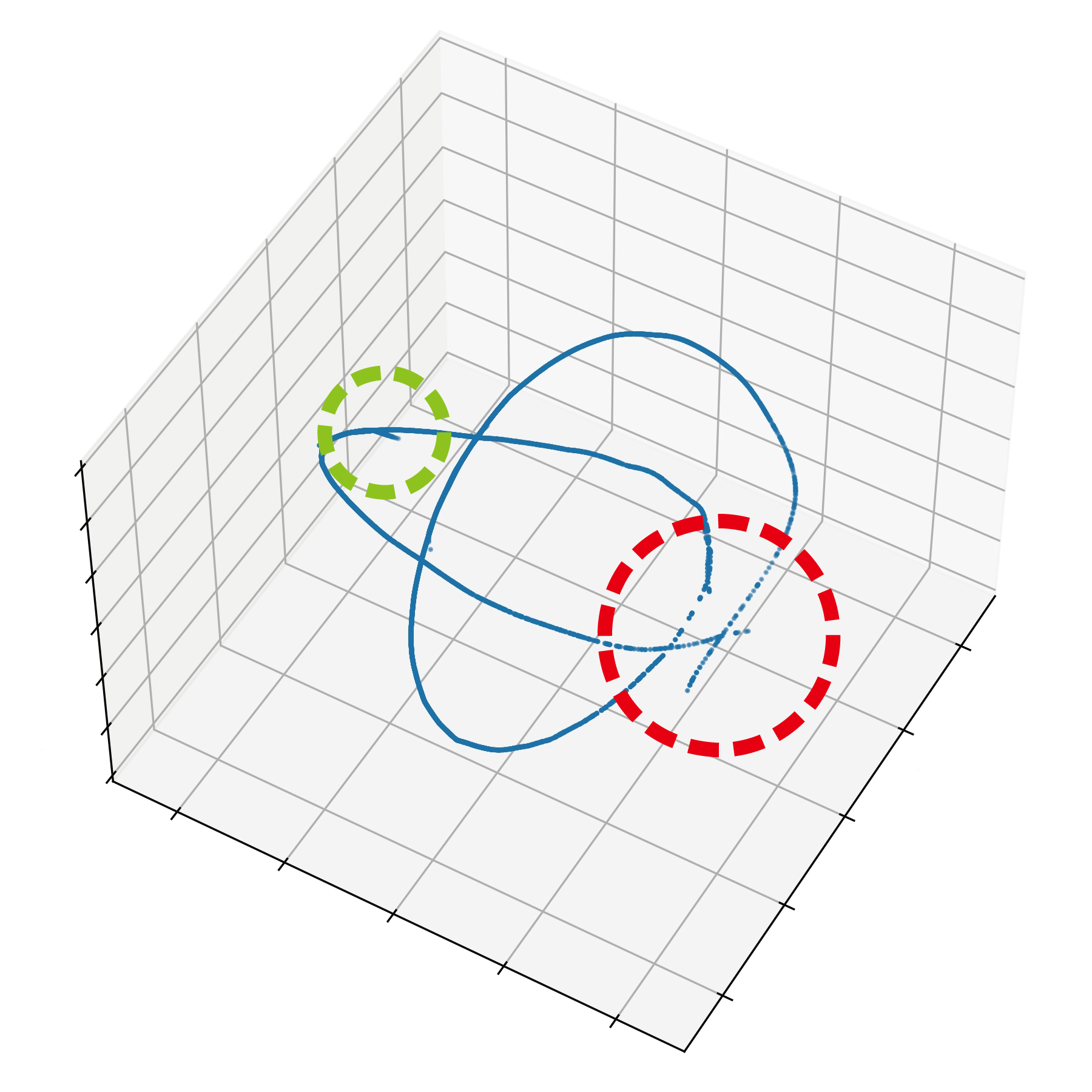}}        
    \caption{}
\end{subfigure}
\caption{(A) The original torus dataset.
(B) Estimation of a torus by using single chart $\mathcal{M}$-flow.
(C) Estimation of density on torus without constraints densities of overlapping regions.
(D) The original Trefoil knot dataset.
(E) Estimation of a trefoil knot by using multiple charts with overlapping regions without constraints.}
\label{motivation}
\end{center}
\vskip -0.2in
\end{figure}

The Flow layer corresponds to a diffeomorphism which is a invertible differentiable function preserving smooth structure.  
In many manifolds, it can be proved mathematically that there are no diffeomorphisms from the manifold to any subset of Euclidian space $\mathbb{R}^n$ where $n$ is the dimension of the manifold .
For instance, there are no diffeomorphisms from the $2-$dimensional torus to any subset of $\mathbb{R}^2$ since their topological structures are different.
Therefore, it can be very difficult for a flow model to learn a whole manifold if the manifold is topologically complicated.
We conducted several experiments of manifold learning using a flow model on the $2-$dimensional torus and confirmed that the quality of sampling using the model was not good. (see Figure \ref{motivation}(b).)

By the definition of a differentiable manifold, there are open subsets $U_i$ of a manifold $\mathcal{M}$ and homeomorphisms $\varphi_i$ from $U_i$ to an open subset of some Euclidean space $\mathbb{R}^n$, such that $\varphi_i \circ \varphi_j^{-1}$ is diffeomorphism for any $i$ and $j$. 
Given this idea, we think it is natural to use multiple flow models for the manifold learning task. 
Multi-chart flows \cite{kalatzis2021multi} uses multiple flow models after dividing data points.
It was confirmed in our individual experiments that using more than one flow model for manifold learning improves the results a lot in the sense that the sampled points usually do not escape largely from the original manifold in our toy example experiments.
However, we also found that if we use partitions that do not allow overlap, the model does not work well, especially on the boundaries of the partitions. (see Figure \ref{motivation}(b).)
In some cases, sampling quality itself is not good, or more commonly, there is a difference between sampled points in other partitions, resulting in the disconnected or not smooth manifold. (see Figure \ref{motivation}(b).) 

We solve this problem using an open cover, a family of open subsets overlapping each other.
We add a loss term to make the parts of a manifold attached together smoothly using the data points on the overlapping regions.
To get an open cover of a manifold, we used Mapper. \cite{singh2007topological}
One can also use soft clustering assignment methods to get an open cover. 
However, additional work is needed such as setting a threshold for probability.

After obtaining an open cover, manifold learning is done for each open set.
We could confirm that allowing overlap between open sets can improve the manifold learning task.
However, if the density estimation is done in each open set, then the density on overlapping regions becomes higher than what it should be (see Figure \ref{motivation}(c)). 
We took the disintegration concept from probability theory so that the probability is well estimated in the overlapping regions also.

\subsection{Mapper}
\label{Mapper}

\textit{Mapper} \cite{singh2007topological} is a tool for the representation of a data manifold as a network while preserving some topological features.
The nodes of a \textit{Mapper} network correspond to a simpler subset of the data manifold and the connection between the nodes describes the global structure of the data manifold.
Based on this property, we developed a method obtaining global geometry and density on the data manifold via local inferences.

We consider a topological manifold $\mathcal{M}$ and a continuous function $h:\mathcal{M} \to \mathbb{R}$ called a \textit{lens function}. 
Given an interval in $h(\mathcal{M}) \subset \mathbb{R}$, we extract a subset of $\mathcal{M}$ whose values for $h$ lie in the interval.
Using a clustering method (e.g., hierarchical linkage clustering, DBSCAN, etc.), this subset is decomposed to form clusters, which represent nodes of the \textit{Mapper} network.
By repeating this process for intervals covering $h(\mathcal{M})$, we can obtain various nodes.
The nodes are connected if corresponding clusters share some data points.

Theoretically, nodes in the Mapper network correspond to critical values if $h$ satisfies some regularity conditions in the sense of Morse Theory \cite{milnor2016morse}.
Moreover, the original manifold $\mathcal{M}$ and the Mapper network are topologically equivalent if a family of clusters forms a good cover \cite{edelsbrunner2010computational}.
For more detailed descriptions, see Appendix \ref{topology}. 

\subsection{Atlas flow}\label{sec:method:atlasflow}

Consider a data-generating process on an $n$-dimensional manifold $\mathcal{M} \subset \mathbb{R}^d$ with probability $p^\mathcal{M}$ on $\mathcal{M}$ and assume that we have a finite sampled dataset with some noise.
We introduce a novel generative model that contains the geometric and topological information of $\mathcal{M}$ and the probabilistic/statistical information of $p^\mathcal{M}$.
We call this model \textit{Atlas flow}.
Figure \ref{method} illustrates framework of our model.

Given a dataset, we infer the shape of $\mathcal{M}$ and the density of $p^\mathcal{M}$ under the assumption that $\mathcal{M}$ is $n$-dimensional compact Riemannian manifold embedded in $\mathbb{R}^d$.

\subsubsection{Manifold learning}

Differential manifold theory ensures that there exists a local coordinate chart system $\{ U_k, \varphi_k\}_{k=1}^L$ of a subset of $\mathbb{R}^d$ such that the \textit{coordinate domains} $\{ U_k \}$ cover $\mathcal{M}$, i.e. 
\begin{equation}
\mathcal{M} \subset \bigcup_{k=1}^L U_k \subset \mathbb{R}^d \;\text{ and }\; \mathcal{M} \cap U_k \neq \phi
\end{equation}
and $\varphi_k : U_k \to V_k \times \tilde{V}_k \subset \mathbb{R}^n \times \mathbb{R}^{d-n}$ is a \textit{coordinate map} with
\begin{equation}
\varphi_k(\mathcal{M} \cap U_k) = V_k \times \{0\}^{d-n}
\end{equation}
for each $k=1,\dots,L$.
(see Appendix \ref{geometry}.)
We note that $\varphi_k$ is a diffeomorphism between coordinate domains $U_k \cap \mathcal{M}$ and a latent space $V_k \subset \mathbb{R}^n$.
Also, $\mathcal{M}$ can be covered by the finite number of coordinate domains since we assume $\mathcal{M}$ is compact.

Each node of the Mapper construction corresponds to a coordinate domain $\{ U_k \}_{k=1}^L$.
Combining all nodes in this correspondence, we get a covering of $\mathcal{M}$.
Then we use a flow-based model to train $\varphi_k$ on $U_k$ in several steps.

Since a flow-based model largely depends on an initial guess, an appropriate initial guess is required to stabilize the model.
We first pre-trained $\varphi_k$ with a dimension reduction method such that the image of $\varphi_k$ is equal to the result of the dimension reduction (Algorithm \ref{alg:pre}).
We found that Isomap fits our flow model well, presumably because Isomap preserves the simple structure of the coordinate domain well.

We set the reconstruction loss for $x \in U_k$ as
\begin{align}
    \mathcal{L}_\text{recon} (x) &= \| x - \mathrm{Recon}_k (x) \|_2 \nonumber\\
    &= \|x - (\varphi_k^{-1} \circ \mathrm{Proj} \circ \varphi_k)(x) \|_2,
\end{align}
where $\mathrm{Proj} : \mathbb{R}^n \times \mathbb{R}^{d-n} \to \mathbb{R}^n \times \{0\}^{d-n}$ is the orthogonal projection map (Algorithm \ref{alg:recon}).
We define the pairwise distance loss defined for a mini-batch $\mathcal{B}$ as
\begin{equation}
    \mathcal{L}_\text{dist} = \mathbb{E} \left[ (D_{ij} - \|x_i - x_j\|_2)^2 \right],
\end{equation}
where $x_i,x_j \in \mathcal{B}$ and $D=[D_{ij}]$ is the pairwise distance matrix whose entries are geodesic distances on the $k$nn graph of Isomap (Algorithm \ref{alg:pdl}).
Then we combined the two losses with loss weight hyperparameter $\lambda$ as follows
\begin{equation}
    \mathcal{L} = \lambda \mathcal{L}_\text{dist} + (1-\lambda) \mathcal{L}_\text{recon}.
\end{equation}

The reason for using pairwise distance is that we noticed that the embedded Isomap result is sometimes far from ideal embedding to the low dimensional space.
This can be caused by the initial guess or by an embedding into a too low-dimensional space.
Since the embedding of Isomap is determined by the pairwise distance, we decided to use the pairwise distance directly.
We also observed that sudden change of the loss may confuse the model, and that's why we slowly decrease the $\lambda$.
With pre-training by Isomap, this process remarkably stabilizes the training process of our flow model.

\begin{figure}
    \centering
    \centerline{\includegraphics[width=0.8\columnwidth]{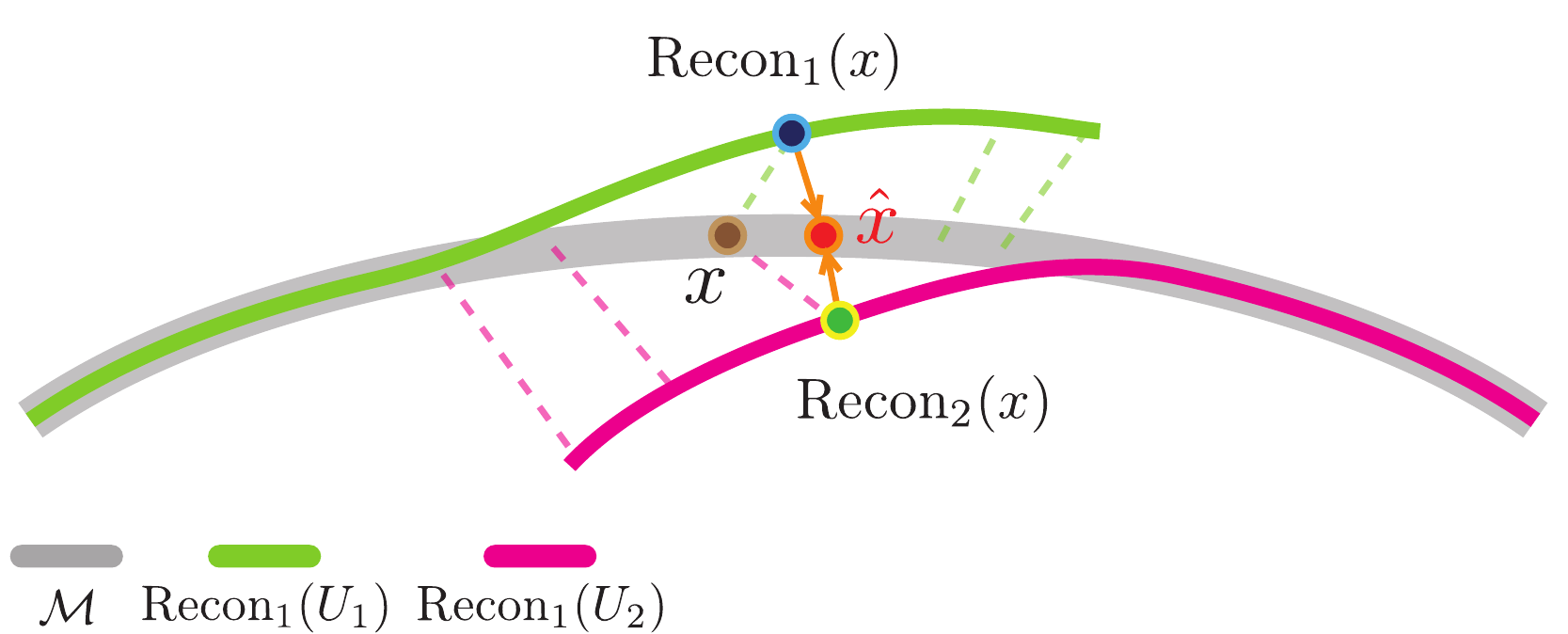}}
    \caption{Reconstruction points $\mathrm{Recon}_k (x)$ are exactly on the reconstructed coordinate domain $\mathrm{Recon}_k(U_k)$. These may not coincide initially. Thus we set expected point $\hat{x}=(\mathrm{Recon}_1(x)+\mathrm{Recon}_2(x))/2$ and adjust coordinate maps $\varphi_k$ so that $\mathrm{Recon}_k(x)=\hat{x}$. }
    \label{method_compaibility}
\end{figure}

We note that coordinate domains may overlap.
In differential manifold theory, compatibility between coordinate maps on the overlapping regions is necessary.
The compatibility makes it possible for local coordinates to extend to the global structure of a manifold.
Thus, it is required to fine-tune $\varphi_k$ for compatibility.
Since data points are sampled on the manifold with noise and $\varphi_k$ is an approximated coordinate map, the reconstructions of a point $x$ by distinct coordinate maps may differ.
We denote the expected point of $x$ by $\hat{x}$, which is obtained as the average of the projections of $x$ to the estimated data manifold in each $U_k$.
In the final step of manifold learning, we adjusted $\varphi_k$ so that reconstructions become equal on the overlapping regions by reducing compatibility loss
\begin{equation}
    \mathcal{L}_\text{comp}(x) = \| x - \hat{x} \|_2
\end{equation}
(See Figure \ref{method_compaibility}, Algorithm \ref{alg:CompLoss}).
The process for compatibility is necessary since the reconstructed coordinate domains may cross each other.
(see Figure \ref{motivation}(e).)

A naive approach for the compatibility loss is adding a loss term based on the distance between the reconstructed sample points.
However, this approach requires heavy memory since more than two flow models should be used and updated at the same time.
Instead, we use only one model while calculating expected points, processing inference, and updating the model. 
Using this method, we can save a lot of memory while calculating compatibility loss.

\subsubsection{Density estimation}

From now, we estimate the density on $\mathcal{M}$ in terms of $p^\mathcal{M}$.
Since we trained the manifold structure by local coordinate charts in the previous subsection, the density should be estimated on each coordinate domain and then combined into the whole density on $\mathcal{M}$.
Thus we needed a appropriate way to decompose density into local densities on coordinate domains.
Decomposition of density function was difficult to apply in practice even if it is well established theoretically. 
Thus we decomposed probability measure instead of density function using disintegration of which concept is based on probability theory\cite{pachl1978disintegration}.
The decomposition by disintegration is a novel approach that has never been before.

\begin{theorem}[Disintegration theorem]
    For a partition $\{ U_{k'}' \}_{k'=1}^{L'}$ of $\mathcal{M}$, the probability $p^\mathcal{M}$ can be decomposed as 
    \begin{equation}
        p^\mathcal{M} = \sum_{k'=1}^{L'} p^{U'}_{k'} \nu(k'),
    \end{equation}
    where $p^{U'}_{k'}$ is a probability on $\mathcal{M}$ with $p^{U'}_{k'} ( U'_{k'} )=1$ and $\nu$ is the push-forward measure of $p^\mathcal{M}$ by the canonical quotient map $\mathcal{M} \to \{1,\dots,L'\}$.
\end{theorem}

Using disintegration theorem, we can decompose $p^\mathcal{M}$ into 
\begin{equation}
    p^\mathcal{M} = \sum_{k=1}^L c_k p^U_k,
\end{equation}
where $c_k$ is a scaling constant and $p^U_k$ is a probability on $\mathcal{M}$ with $p^U_k(U_k)=1$.

We state more details of disintegration in Appendix \ref{probability}.

For each $k=1,\dots,L$, consider learnable diffeomorphisms $\gamma_k : V_k \to W_k$ called density transformation map, where $W_k \subset \mathbb{R}^n$ adopts probability $p^W_k$ according to the standard normal distribution, i.e. another latent space with tractable density.
We then train $\gamma_k$ by negative log-likelihood loss $\mathcal{L}_\text{density}$ of $p^W_k$.
The value of $p^W_k$ at the latent variable from the sampled data point is calculated from the following formula (Algorithm \ref{alg:dll}).
For $x \in U_k$,
\begin{align}
p^U_k (x)
&= p^V_k (\varphi_k(x)) \left| \det [J_{\varphi_k}^T (\varphi_k(x)) J_{\varphi_k} (\varphi_k(x)) ] \right|^{\frac{1}{2}} \nonumber \\
&= p^W_k (\gamma_k(\varphi_k (x))) \left| \det J_{\gamma_k} (\gamma_k(\varphi_k(x))) \right| \nonumber \\
&\qquad \times \left| \det [J_{\varphi_k}^T (\varphi_k(x)) J_{\varphi_k} (\varphi_k(x)) ] \right|^{\frac{1}{2}}, \label{p_cal}
\end{align}
where $J_*$ is the Jacobian matrix of a function $*$.

The last term of (\ref{p_cal}) only depends on the coordinate map, not on the density transforming map.
Calculating the last term of (\ref{p_cal}) is also heavy, so we omit that term in (Algorithm \ref{alg:dll}).

We first trained $p^U_k$ by data samples in $U_k$.
However, this process provoked serious problem that the density on the overlapping regions becomes higher since the densities are also overlapped.
Therefore we noticed that each $p^U_k$ should have a lower density in the overlapping regions in $U_k$ by the construction of $p^U_k$.
(Appendix \ref{probability}.)
In order to consider this property to learning $p^U_k$, we used a special batch technique.
We denote by $m_x$ such that $x$ is contained in $m_x$ coordinate domains and let $m_k$ be the least common multiplier of $\{ m_x : x \in U_k \}$.
Then we made mini-batches for training $\gamma_k$ by bootstrapping according to a density on $x \in U_k$ as $m_x/m_k$. 
This batch technique allows that the density of $p^U_k$ gets lower in the overlapping region inversely proportional to the number of overlapping coordinate domains.

\section{Experiment}\label{sec:exp}
We used our method on some datasets, ranging from the well-known manifolds such as trefoil knot and torus in $\mathbb{R}^3$, to a not well-known space such as $\mathcal{W}$ space of StyleGAN2.
For the well-known manifolds, we consider generated sample points to evaluate the performance.
For the $\mathcal{W}$ space, we sampled points on it and generated images from those points.
Furthermore, we demonstrated that the compatibility of coordinate maps is well considered in our Atlas flow model from the observation that overlapping coordinate domains are topologically the same.

Every flow model in our experiments is based on rational-quadratic splines.
The number of layers depends on the datasets. 
Usually, we choose $11-13$ layers per one flow model. For more experimental detail, please refer to Appendix \ref{sec:exp_detail}.

\subsection{Trefoil knot}

First, We want to apply the method to complicated $1$-manifold in $\mathbb{R}^3$, trefoil knot with noise.
Obviously, this manifold cannot be covered by a single chart.
Therefore, using several coordinate domains is meaningful and necessary.
From the Mapper algorithm, we can get four coordinate domains on the trefoil knot.
We use 11 coupling layers for manifold learning, and 11 autoregressive layers\cite{huang2018neural} for density estimation.

The overlapping regions are well attached to each other thanks to compatibility loss, and density is also well estimated.

\begin{figure}[ht]
\vskip 0in
\begin{center}
    \begin{subfigure}[b]{0.19\columnwidth}
        \centering
        \centerline{\includegraphics[width=\columnwidth]{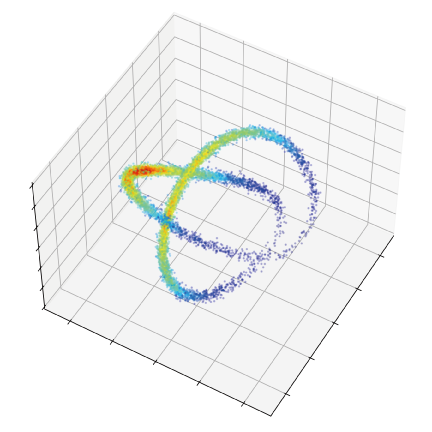}}
        \caption{}
    \end{subfigure}\quad
    \begin{subfigure}[b]{0.57\columnwidth}
        \centering
        \centerline{\includegraphics[width=\columnwidth]{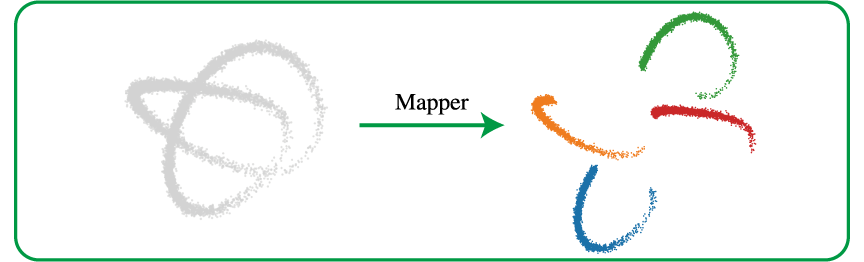}}
        \caption{}
    \end{subfigure}\quad
    \begin{subfigure}[b]{0.19\columnwidth}
        \centering
        \centerline{\includegraphics[width=\columnwidth]{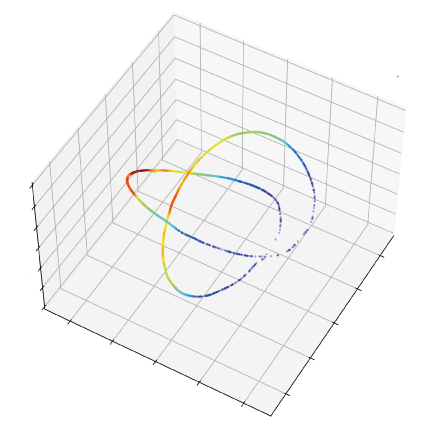}}
        \caption{}
    \end{subfigure}
\caption{(A) Synthetic dataset. The colors of points represent densities estimated by the KDE algorithm. 
(B) Four coordinate domains. 
(C) Sampling points by Atlas flow model. The colors of points represent densities estimated by the KDE algorithm.}
\label{knot}
\end{center}
\end{figure}

\subsection{Torus}

\begin{figure}[ht]
\vskip 0in
\begin{center}
\begin{subfigure}[b]{0.19\columnwidth}
        \centering
        \centerline{\includegraphics[width=\columnwidth]{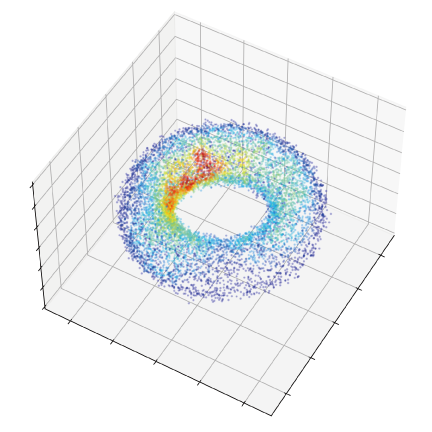}}
        \caption{}
    \end{subfigure}\quad
    \begin{subfigure}[b]{0.57\columnwidth}
        \centering
        \centerline{\includegraphics[width=\columnwidth]{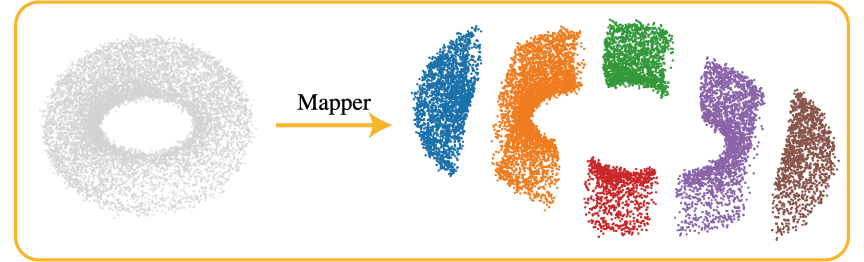}}        
        \caption{}
    \end{subfigure}\quad
    \begin{subfigure}[b]{0.19\columnwidth}
        \centering
        \centerline{\includegraphics[width=\columnwidth]{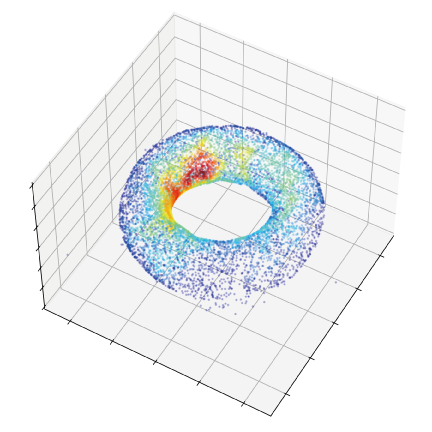}}
        \caption{}
    \end{subfigure}
\caption{(A) Synthetic dataset. The colors of points represent densities estimated by the KDE algorithm. 
(B) Six coordinate domains. 
(C) Sampling points by Atlas flow model. The colors of points represent densities estimated by the KDE algorithm.}
\label{torus}
\end{center}
\end{figure}

For a more complicated manifold, we apply the Atlas flow to the $2-$dimensional torus in $\mathbb{R}^3$ with noise.
Getting six coordinate domains from Mapper, we trained the model for each cover.
We use 13 coupling layers both for the manifold learning and the density estimation.
The density functions on the overlapping regions are well estimated in this experiment.
This experiment shows that our density adjustment technique is well applied.

\subsection{StyleGAN2}

\begin{wrapfigure}{r}{0.5\textwidth} 
    \centering
    \centerline{\includegraphics[width=0.5\columnwidth]{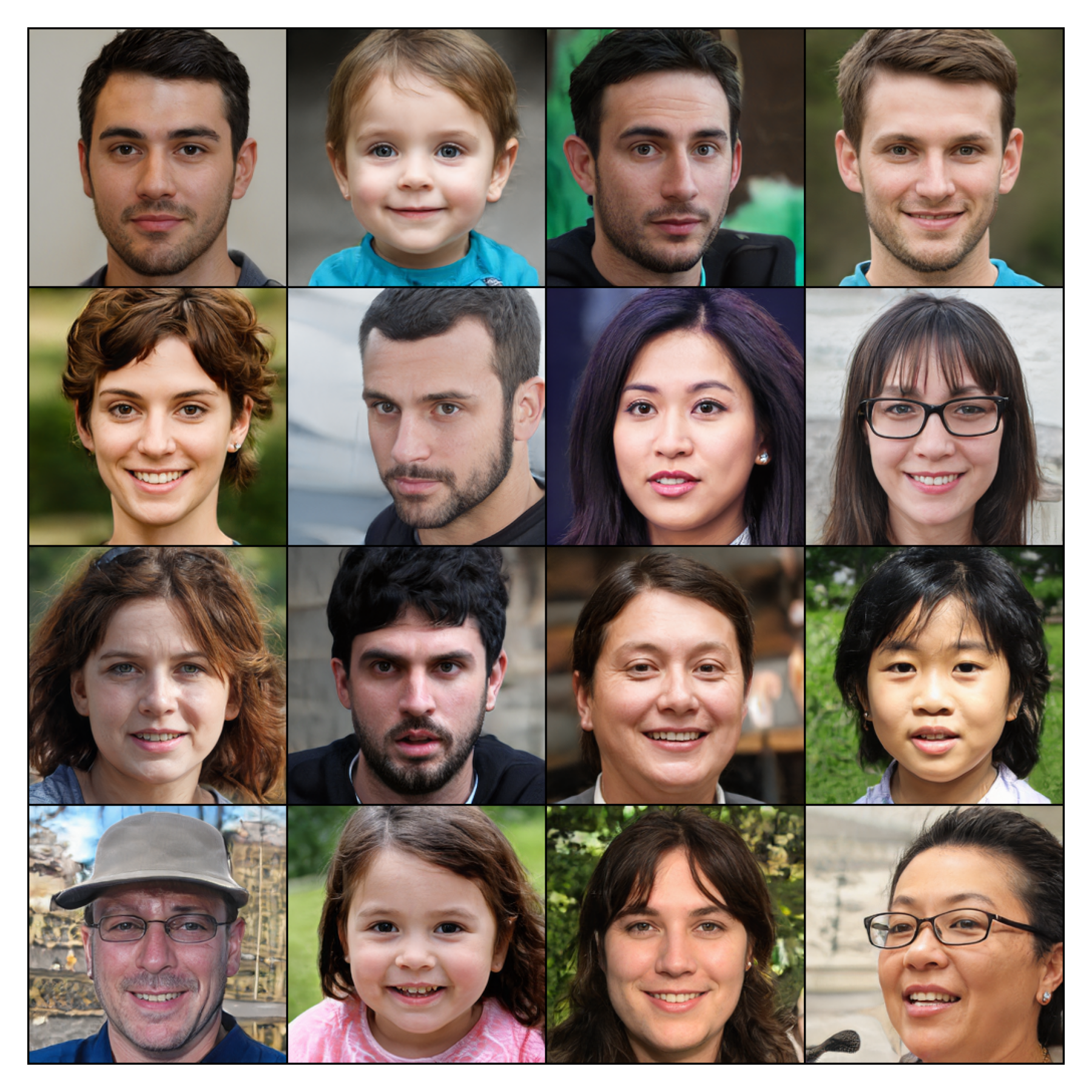}}
    \caption{Generated images from a submanifold in the $\mathcal{W}$ space.}
    \label{stylegan2:sphere}
\end{wrapfigure}
We analyze the style vector space $\mathcal{W}$ of StyleGAN2 \cite{karras2020analyzing} using the pre-trained model on the FFHQ dataset.
In the StyleGAN2 model, the style vector in $\mathcal{W}$ can be obtained using the mapping network of StyleGAN2 from the standard normal distribution in $\mathbb{R}^{512}$, the $\mathcal{Z}$ space.

The next step of the generator is obtaining the image using the synthesis network of the model from the style vector.
In our experiment, we noticed that the generated image is mainly determined by the direction of the vector in $\mathcal{Z}$ space.
Hence, to analyze the $\mathcal{W}$ space, we took $4 \times 10^4$ samples from $S^{511}\subset\mathbb{R}^{512}$ on the $\mathcal{Z}$ space, which represents every direction of $\mathbb{R}^{512}$, and observed the data points which come from the synthesis network of the generator model.

From a geometric viewpoint, the manifold can be defined as a collection of coordinate charts and transition maps between the coordinate charts.
Using our method, we present a new geometric method to analyze the $\mathcal{W}$ space by training coordinate maps from the manifold.
We checked that the quality of images from our generating model is plausible (Figure \ref{stylegan2:sphere}).

Next, we show that the local coordinate systems in the overlapping region between coordinate domains are well matched. 
In this experiment,  we first fixed a vector from the overlapping region in $\mathcal{W}$ space. 
Using a coordinate map and a density transforming map, we can perturb the vector in the $W$ space for each coordinate chart.
Since the coordinate domains cover a small neighborhood of the vector, the images generated from the neighborhood should match each other even though the scaling and direction of the axis can be different and skewed.

\begin{figure*}[ht]
\vskip 0in
\begin{center}
\centerline{\includegraphics[width=\textwidth]{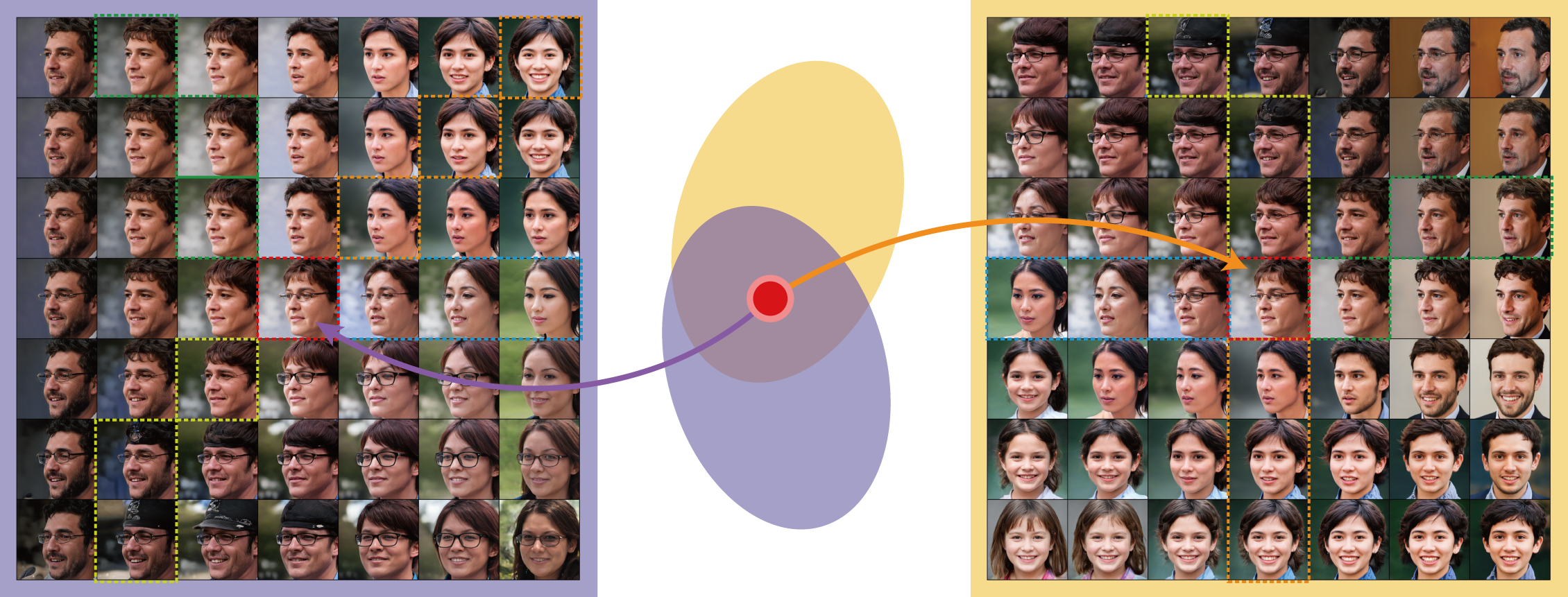}}
\caption{Generated images from perturbation vectors in each coordinate domain. The Center image with a red dotted border is from the fixed vector on the overlapping region. The images with the other color dotted borders represent different coordinate axis on two coordinate domains.}
\label{perturbation_on_s2}
\end{center}
\vskip -0.2in
\end{figure*}

For the experiment, we first sampled from $S^2$ using the first $3$ coordinates of $\mathcal{Z}$ space and multiplied an orthogonal matrix of size $512 \times 512$ so that various directions on $\mathbb{R}^{512}$ can be regarded while it preserves the dimension of the manifold.
Figure \ref{perturbation_on_s2} illustrates the images generated from the vectors in the perturbation neighborhood in each coordinate domain.
From the result, we demonstrate the fact that the small neighborhood of the vector can be matched up to a skewed axis and different scales.

In our experiments using the StyleGAN2 model, we use 13 coupling layers both for the manifold learning and the density estimation after getting two coordinate domains from the Mapper algorithm.

\subsection{Comparing with the partitions} \label{exp:cover_parti}

We designed an experiment that shows the advantage of using a cover instead of partition\cite{kalatzis2021multi} in the training process of the Atlas flow. 

The experiment shows the reconstructed data points of boundaries of partition/cover of the manifold.
We sampled $10,000$ points of the torus with noise in $\mathbb{R}^3$.
We define a data label for each data point by a partition/cover label in which the point is contained.

\begin{figure}[ht]
\vskip 0in
\begin{center}
\begin{subfigure}[b]{0.48\columnwidth}
    \centering
    \centerline{\includegraphics[width=\columnwidth]{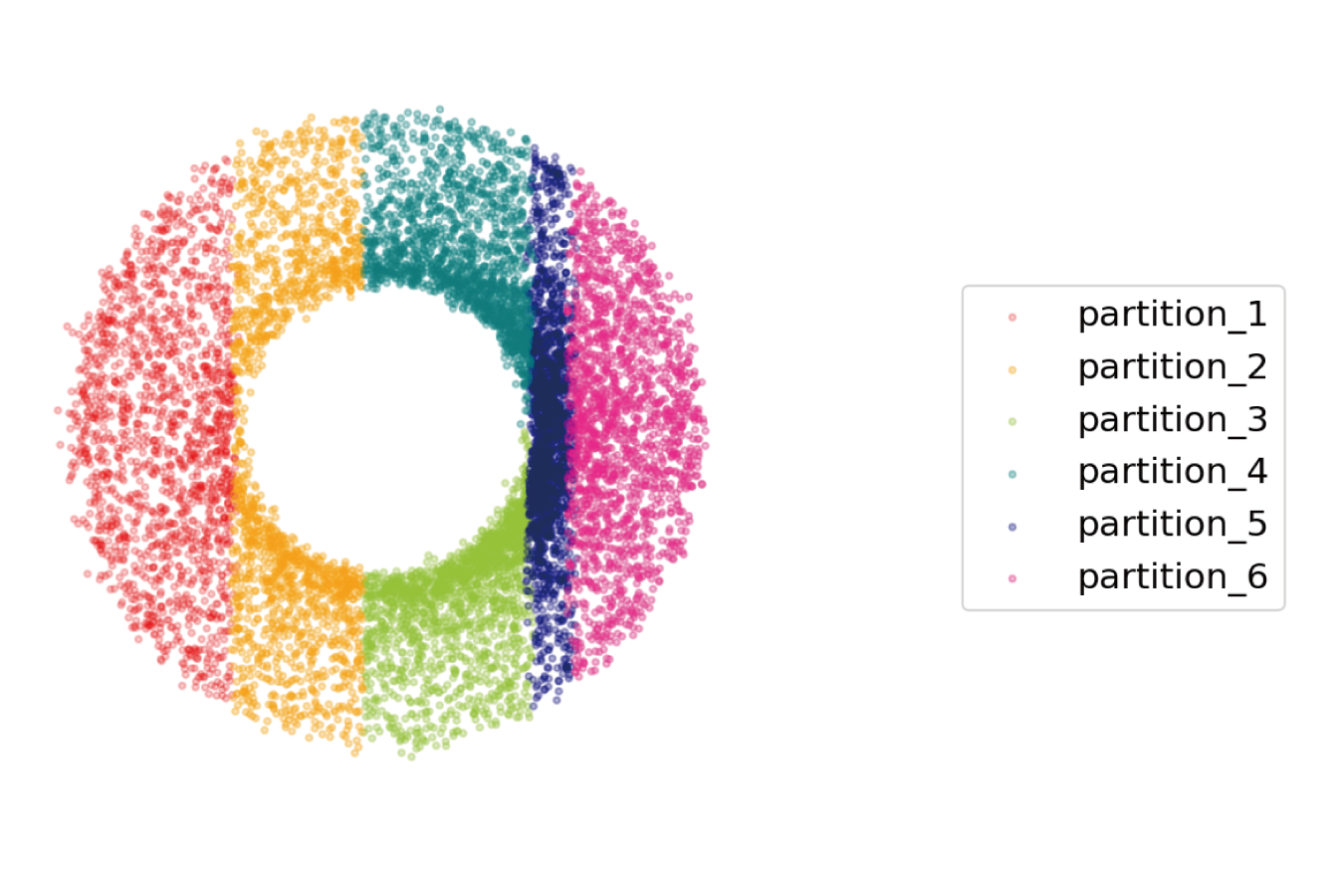}}
    \caption{}
\end{subfigure}
\begin{subfigure}[b]{0.48\columnwidth}
    \centering
    \centerline{\includegraphics[width=\columnwidth]{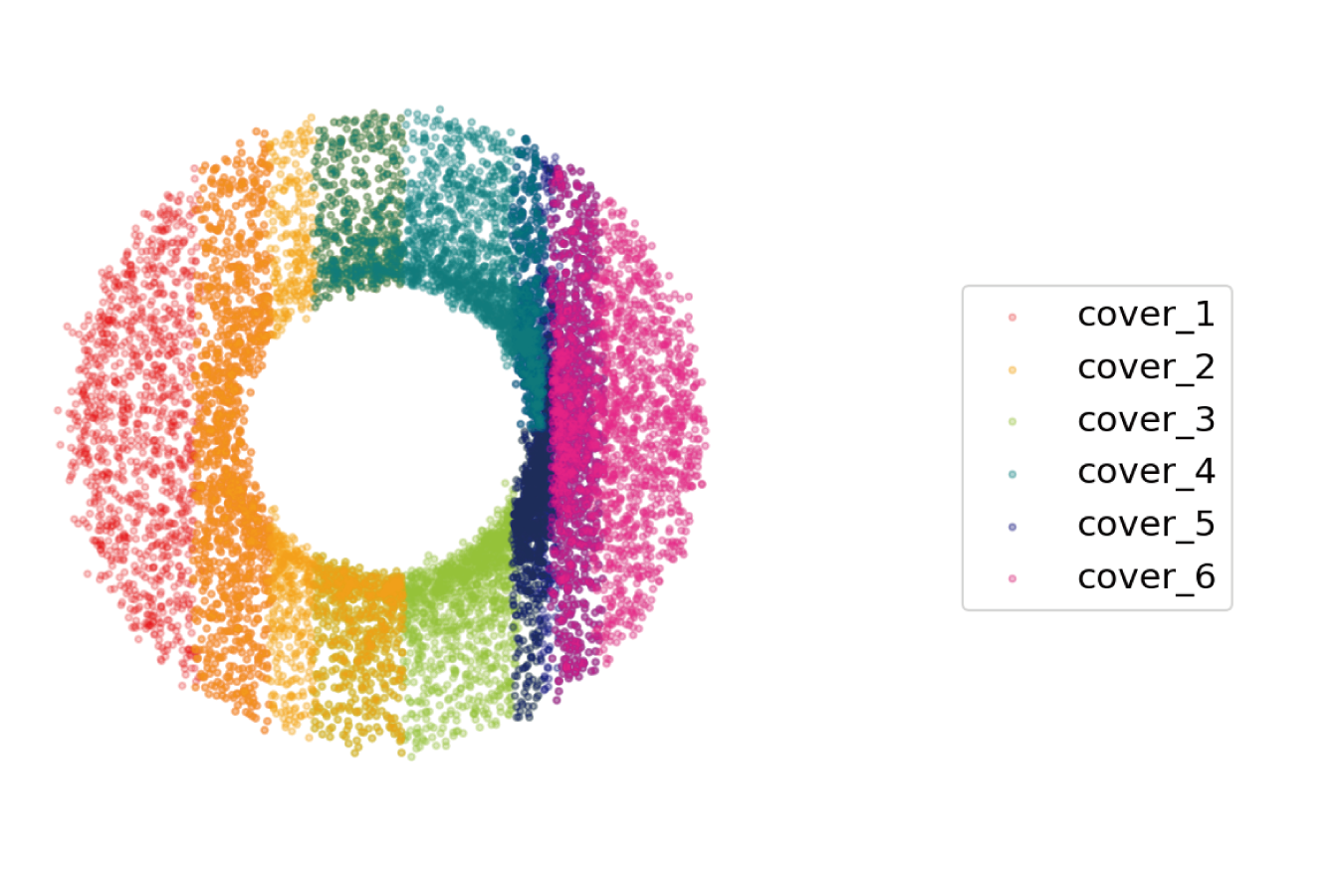}}
    \caption{}
\end{subfigure}
\caption{(A) is the partition label and (B) is the cover label.
}
\label{data_label}
\end{center}
\end{figure}

The experiment process is following:
\begin{enumerate}
    \item The data label is illustrated in Figure \ref{data_label}. 
    \item We took data points in $30 \%$ bands of overlapping regions around the boundaries from partition.
    \item We used models which are trained by using partition-wise and cover-wise datasets.  
    For fairness of comparison, the models were trained by the same hyper-parameters.
    \item For the points on the boundary of the cover/partition, reconstruction was performed with a model trained, not on the cover containing the points, but on the cover adjacent to the cover containing the points.
    \item Compute and compare the MSE losses between the original data and its reconstructions by models from partition and cover.
\end{enumerate}

\begin{figure}[ht]
\vskip 0in
\begin{center}
\begin{subfigure}[b]{0.32\columnwidth}
    \centering
    \centerline{\includegraphics[width=\columnwidth]{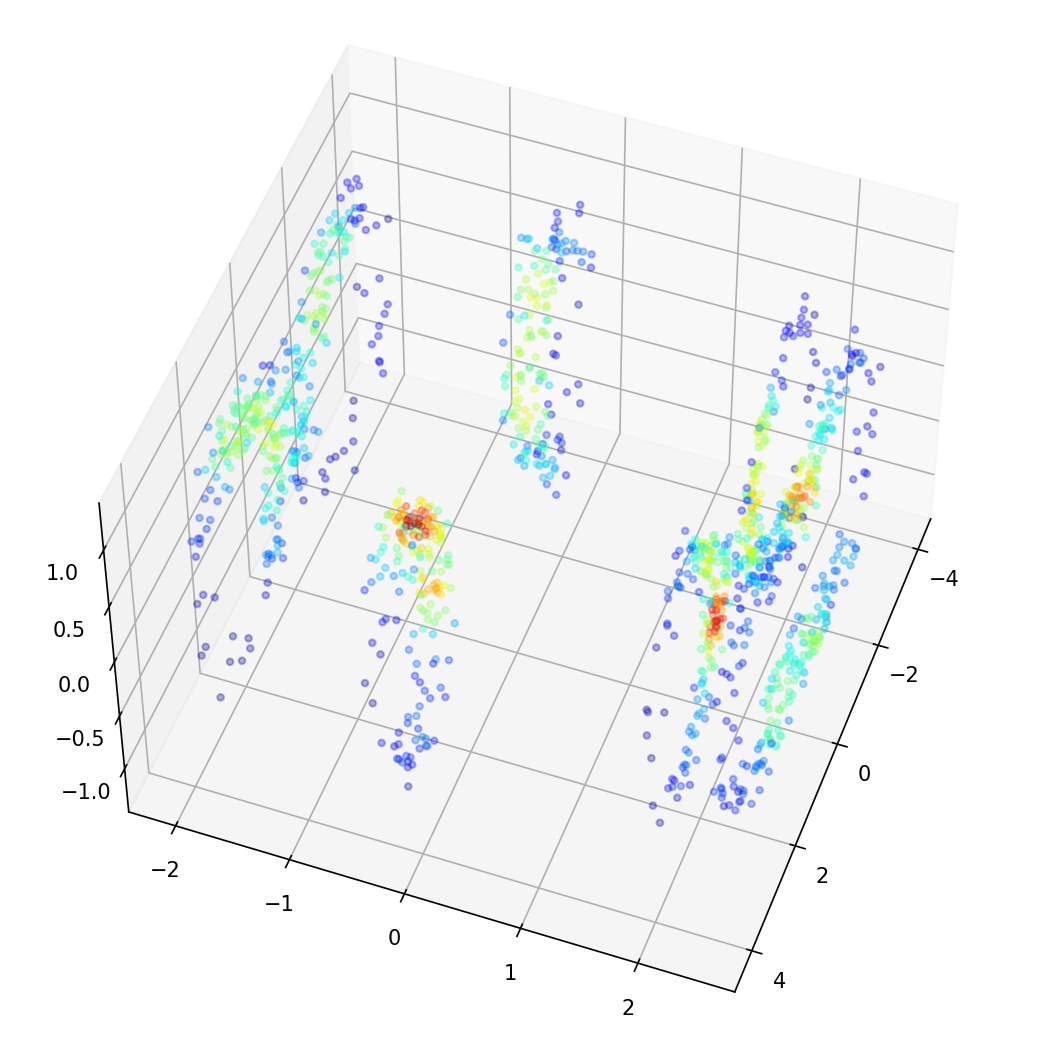}}
    \caption{}
\end{subfigure}
\begin{subfigure}[b]{0.32\columnwidth}
    \centering
    \centerline{\includegraphics[width=\columnwidth]{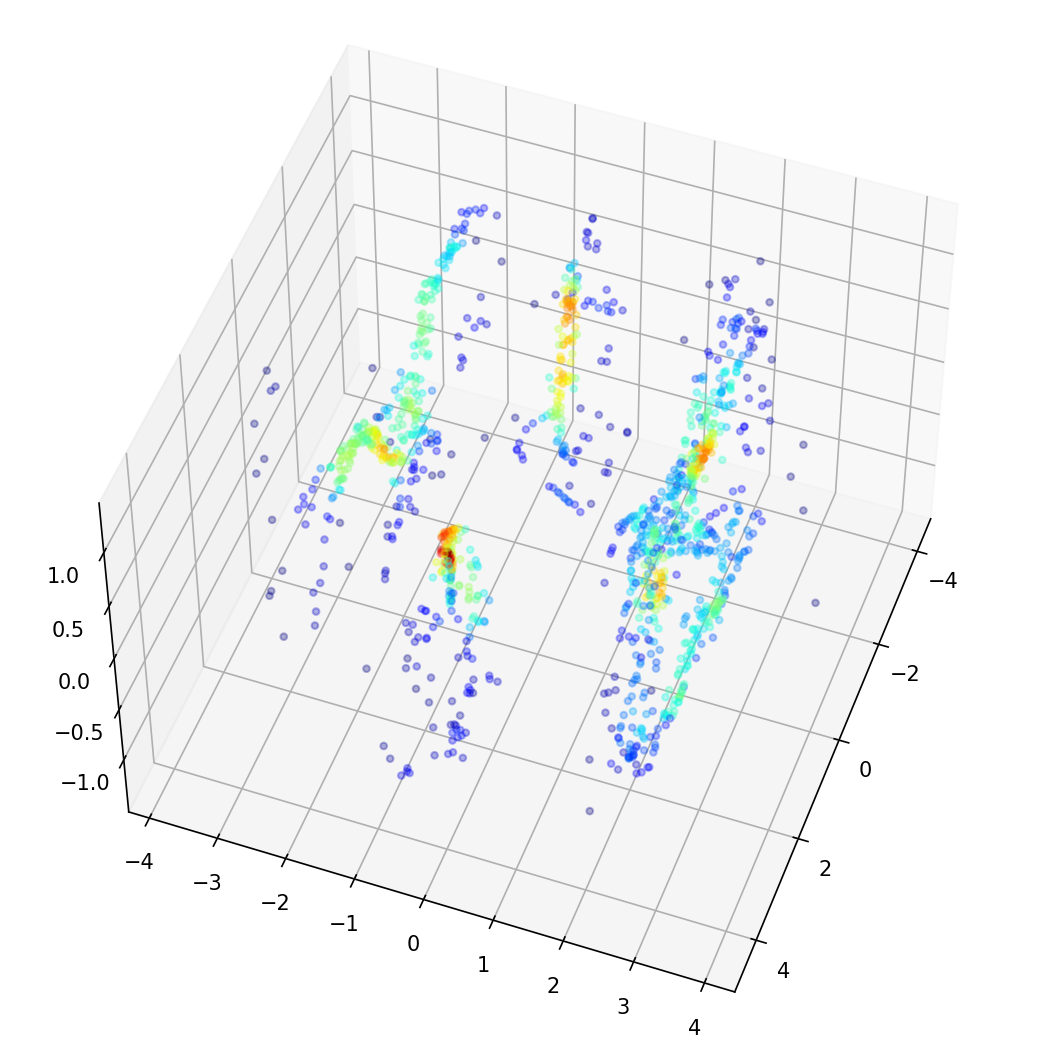}}
    \caption{}
\end{subfigure}
\begin{subfigure}[b]{0.32\columnwidth}
    \centering
    \centerline{\includegraphics[width=\columnwidth]{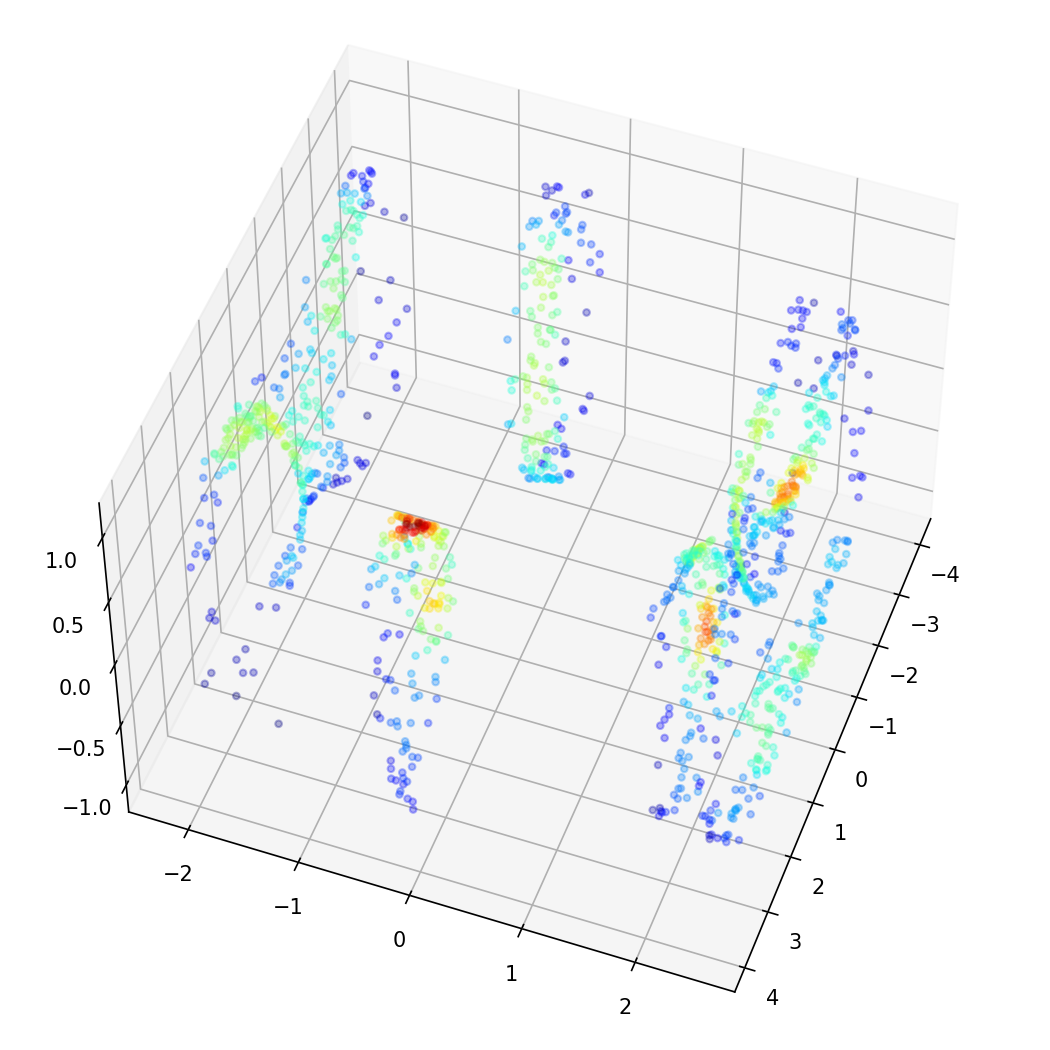}}
    \caption{}
\end{subfigure}
\caption{(A) The boundary points in the original torus dataset.
(B) The reconstructed points that is estimated by using models trained by partition.
(C) The reconstructed points that is estimated by using models trained by cover.}
\label{recon_err_image}
\end{center}
\end{figure}

Through the quantitative (See the Tables \ref{recon_err_total} and \ref{recon_err_detail}) and qualitative (See the Figure \ref{recon_err_image}) results of this experiment, we justified that the model from cover is better than the model from partition in the training process of Atlas flow.

\begin{table}[ht]
    \centering
    \begin{tabular}{c|c}
        \hline\hline
         & Average Reconstruction Errors \\
         \hline\hline
        Partition & 0.1176 \\
        \hline
        Covers & \textbf{0.0181} \\
        \hline\hline
    \end{tabular}
    \caption{
    The table is the average reconstruction error of points in the boundaries.
    $1,344$ points is selected out of $10,000$ data points.}
    \label{recon_err_total}
\end{table}

\begin{table}[ht]
    \centering
    \begin{tabular}{c|cccccc}
        \hline\hline
        (Data label, Model label) &  $(1,2)$ &  $(2,3)$ &  $(2,4)$ &  $(3,5)$ &  $(4,5)$ &  $(5,6)$  \\
        \hline
        Partitions &  0.1011 &  0.1147 &  0.0287 &  0.0507 &  0.0329 &  0.0803 \\
        \hline
        Covers &  \textbf{0.0198} &  \textbf{0.0200} &  \textbf{0.0245} &  \textbf{0.0209} &  \textbf{0.0129} &  \textbf{0.0239} \\
        \hline
        Number of points &  168 &  89 &  79 &  93 &  75 &  168 \\
        \hline\hline
        (Data label, Model label) &  $(2,1)$ &  $(3,2)$ &  $(4,2)$ &  $(5,3)$ &  $(5,4)$ &  $(6,5)$  \\
        \hline
        Partitions &  0.3283 &  0.1905 &  0.1689 &  0.0902 &  0.0453 &  0.0657 \\
        \hline
        Covers &  \textbf{0.0138} &  \textbf{0.0175} &  \textbf{0.0187} &  \textbf{0.0111} &  \textbf{0.0128} &  \textbf{0.0183} \\
        \hline
        Number of points &  168 &  94 &  74 &  84 &  84 &  168 \\
        \hline\hline
    \end{tabular}
    \vskip 0.2in
    \caption{For each pair of data label and model label, the model from cover is better than the model from partition.}
    \label{recon_err_detail}
\end{table}

\newpage
\subsection{Comparing with the single chart}

\begin{figure}[ht]
    \centering
    \includegraphics[width =0.7\columnwidth]{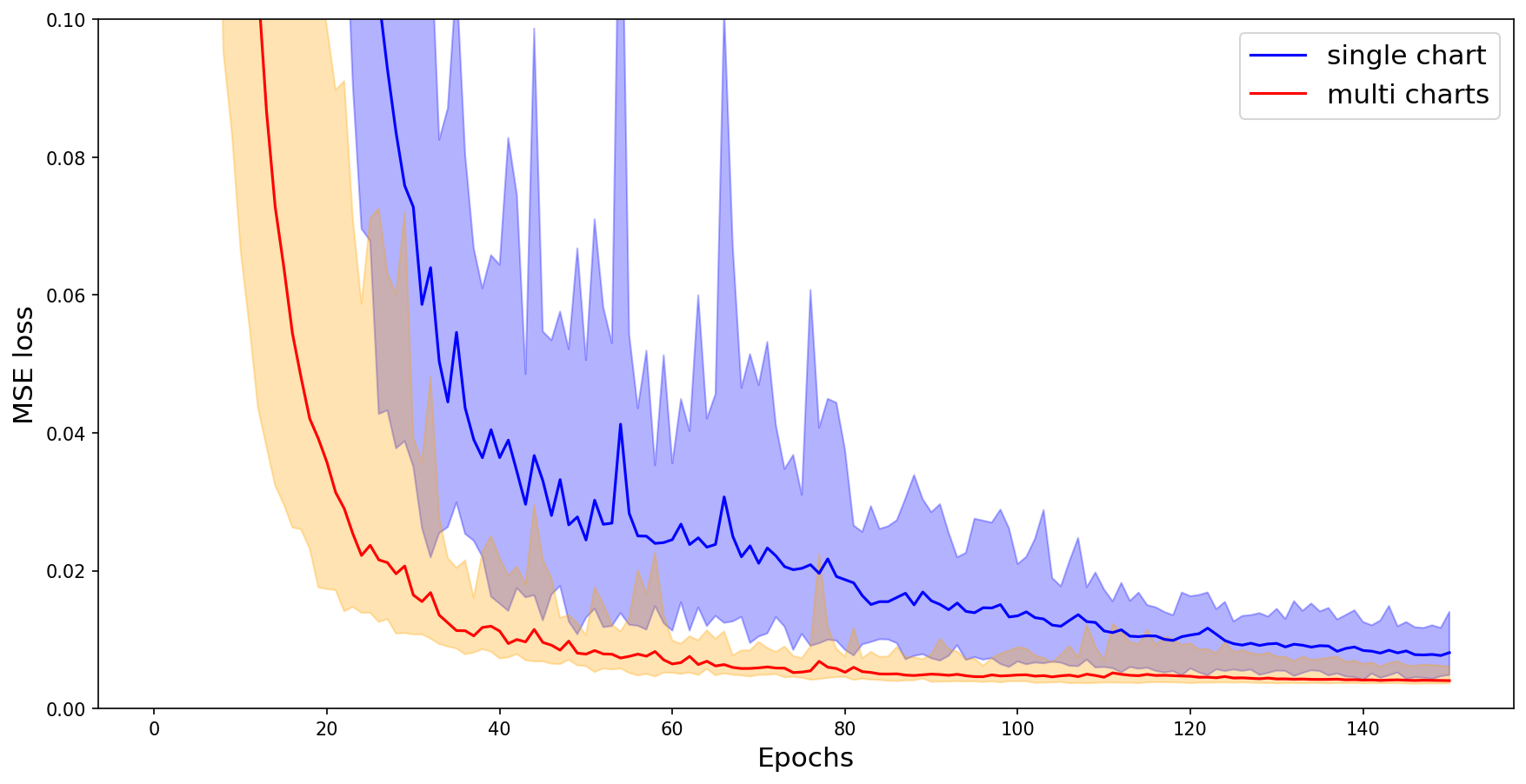}
    \caption{The light blue region shows the minimum/maximum of the MSE loss for each epoch when we used a single chart, and the blue line means the average of the loss for each epoch.
    Similarly, we presented the MSE loss for each epoch using a light orange region and red line when we used the chart.}
    \label{compare:single}
\end{figure}

We designed an experiment to identify the difference in reconstruction error between a single chart\cite{brehmer2020flows} and multi charts by the cover. 
The $x$-axis of Figure \ref{compare:single} means the number of epochs as the training process progress, and the $y$-axis means the average of MSE loss on the training set.
We ran exactly the same $10$ experiments using noisy torus data and briefly presented them in Figure \ref{compare:single}.

In the training process, we confirmed that the MSE loss decreased more rapidly on the model from cover.
Furthermore, we found that after MSE losses converge, the final loss is lower on the model from cover.

\section{Conclusion}
In this study, we mainly focused on the reconstructibility of a manifold with a possibly complicated topological structure.
Thanks to the development of flow-based models, we can take concepts from differential manifold theory, and use them in machine learning theory.
A key point of our Atlas flow model is that we used local coordinate charts that are compatible with each other.
Only by using the local coordinate charts can the complex structure of a manifold be accurately identified. Compatibility between the local charts should be considered to extend local structures to the a global structure.

The structure of a manifold $\mathcal{M}$ is determined by local coordinate charts.
A coordinate map identifies the corresponding coordinate domain with Euclidean space.
That is, it allows us to regard a manifold as a union of Euclidean spaces which has the advantage of using a coordinate system on the manifold.
Although coordinates do not coincide on overlapping regions, compatibility between coordinate maps ensures the local coordinate systems are patched together to form a global structure.
To ensure the existence of a coordinate system, the coordinate domain should be open subsets in Euclidean space.

One limitation of the flow-based generative model is that the training process is unstable.
In our Atlas flow model, the training process becomes more stable by using simple coordinate domains.
However, it is not enough so we added some steps to stabilize the training process.
For this reason, the training process becomes heavy and slow.
We intend to overcome this limitation by improving the flow model in future work.

In topological data analysis, it is a crucial problem to take a cover for a given manifold.
A cover consisting of simple regions is ideal but it is almost impossible to take such a cover without any information of the manifold.
We use Mapper to overcome this problem.
However, even though a cover from Mapper has a big advantage in that it gives topological information of the manifold, it mainly depends on the lens function and cannot guarantee that it is good enough.
Thus we expect that further research on how to take an ideal cover enhances our Atlas flow model.

%%%%%%%%%%% Reference
\newpage

% \nocite{langley00}

\bibliographystyle{unsrt}
\bibliography{main}

%% appendix
\appendix
\onecolumn

\section{Appendix}

\subsection{Topological background} \label{topology}

\textit{Mapper} \cite{singh2007topological} is the most useful method for dimension reduction and exploratory data analysis based on the topological data analysis (TDA) \cite{nicolau2011topology, li2015identification}.
The result of Mapper represents a dataset as a network, which preserve the topological features of the original data point cloud.
Using dimension reduction (for a filter function) and clustering, the resulting network captures the local and global structure of the dataset simultaneously.
More precisely, the node of the network corresponds to a connected component of some level set which is the local simplest region of a data manifold.
The edges connecting these region shows the global feature of a data manifold, such as topological "holes" (homological feature, precisely) or "shape".
These advantages of Mapper are why we applied Mapper to take a local chart.

In our theoretical view, a bundle of local information of data manifold can be expanded to the global information and these processes will give more precise information of data manifold than trying to take information on whole data manifold at once.

Theoretically, Mapper can be interpreted as two mathematical objects.
\begin{itemize}
    \item A discrete version of Reeb graph
    \item An 1-skeleton of a nerve of a refined pull-back cover.
\end{itemize}
The rest of this section gives some theoretical meaning of Mapper with some regularity assumptions.

\subsubsection{Reeb graph}

Given a topological manifold $\mathcal{M}$ and a continuous function $h:\mathcal{M} \to \mathbb{R}$, the \textit{Reeb graph} $R_h (\mathcal{M})$ of $\mathcal{M}$ with $h$ is the quotient space $\mathcal{M}/\sim$ such that for $x,y \in \mathcal{M}$, $x \sim y$ if and only if they are contained in the same connected component of some level set $h^{-1}(a)$ \cite{shinagawa1991surface}.
We note that strictly speaking, a Reeb graph is not a graph object in a mathematical sense but it can be converted to a graph structure.
Furthermore, we think of a Reeb graph as a directed graph by assigning a direction to each edge associated with the gradient descent direction of $h$.
With some regularity conditions, thanks to the Morse theory \cite{milnor2016morse}, we can reconstruct $\mathcal{M}$ from $R_h(\mathcal{M})$ up to topological equivalence.
That is, the Reeb graph $R_h(\mathcal{M})$ contains the topological information of the original space $\mathcal{M}$ even though $R_h(\mathcal{M})$ is the simpler object.

\subsubsection{Nerve}

Consider a topological sapce $\mathcal{M}$ and a cover $\mathcal{U} = \{U_1, \dots, U_L\}$ on $\mathcal{M}$.
The \textit{nerve} $\mathcal{N}(\mathcal{U})$ of $\mathcal{U}$ is a simplicial complex consisting of all simplex of the form $\{ U_{i_1}, \dots U_{i_k} \}$ such that $U_{i_1} \cap \cdots \cap U_{i_k} \neq \phi$ \cite{edelsbrunner2010computational}.
For a simplex $\{ U_{i_1}, \dots U_{i_k} \}$, $k$ is a dimension of the simplex.
The \textit{1-skeleton} of $\mathcal{N}(\mathcal{U})$ is the subcomplex of $\mathcal{N}(\mathcal{U})$ consisting of simplex of dimension $\leq 1$.
We note that any 1-skeleton is a graph.

From Mapper section in the main context, we call pre-images of intervals $\{ h^{-1}(I) | I \subset h(\mathcal{M}) \}$ \textit{a pull-back cover} and $\{ U_i \}_{k=1}^L$ \textit{a refined pull-back cover}.
Then, by definition, Mapper is equal to a nerve of an 1-skeleton of a nerve of a refined pull-back cover.

A nerve is a useful object in topology and geometry fields and its importance is guaranteed by the nerve theorem \cite{ghrist2014elementary}.

\begin{theorem}[Nerve theorem]
If $\mathcal{U}$ is a good cover, that is, any intersection $U_{i_1} \cap \cdots \cap U_{i_k}$ of elements $\mathcal{U}$ has no holes (contractible), then $\mathcal{N}(\mathcal{U})$ is topologically equivalent to $\mathcal{M}$.
(Precisely, homotopic equivalent.)
\end{theorem}
We do not give a detailed proof or description of mathematical terminologies, such as good cover or homotopic equivalent.
(For more details, see \cite{MR1867354}.) 
However, the nerve theorem means that if we choose an appropriate cover for a data manifold, then Mapper is the most useful method for visualizing and analyzing the topological aspect of the ambient data space.

\subsection{Geometric background} \label{geometry}
\subsubsection{Manifold}
A \textit{$n$-dimensional differential manifold} $\mathcal{M}$ is a topological space adapting differential structure on it \cite{lee2013smooth}.
One important property of $\mathcal{M}$ is the locally Euclidean, that is, $\mathcal{M}$ is covered by open subsets, which are the same to Euclidean space.
More precisely, the differential structure of $\mathcal{M}$ consists of open subsets $ \{ U_k \}$ and invertible maps $\{ \varphi_k : U_k \to \mathbb{R}^n \}$ with following properties.
\begin{itemize}
    \item $\{ U_k \}$ covers $\mathcal{M}$, i.e. $\mathcal{M} = \cup_{k} U_k$.
    \item $\varphi_k$ is a homeomorphism, i.e. the domain and image of $\varphi_k$ have the exectly same in the topological sense.
    \item For any $(U_k,\varphi_k)$ and $(U_l,\varphi_l)$, the transition map $\varphi_l \circ \varphi_k^{-1} : \varphi_k (U_k \cap U_l) \to \varphi_l (U_k \cap U_l)$ is a diffeomorphism.
\end{itemize}
We call such a pair $(U_k,\varphi_k)$ as a \textit{local coordinate chart}.
We note that if $\mathcal{M}$ is compact, then the structure of $\mathcal{M}$ can consist of the finite number of charts.

A submanifold of $\mathcal{M}$ is the topological subspace of $\mathcal{M}$ with differential structure induced by the restriction.
Thus we can regard a submanifold as a manifold itself.
If we consider a manifold $\mathcal{M}$ as a subamanifold of $\mathbb{R}^d$, then we may think that each $\varphi_k$ is a restriction of a diffeomorphism on $\mathbb{R}^d$.

\subsubsection{Rank theorem}
A differential structure of a manifold has special property that charts are not uniquely determined and geometric properties are invariant to choices of charts if they are compatible.
Moreover, we can take charts so that induced submanifold structure become simpler in view of the chart map.
This property is summarized by following theorem.

\begin{theorem}(Submanifold property)
    For a $n$-dimensional submanifold $\mathcal{M}$ in $\mathbb{R}^d$, there are coordinate charts $\{ (U_k,\varphi_k) \}$ of $\mathbb{R}^d$ such that 
    \begin{equation}
        \varphi_k(\mathcal{M} \cap U_k) \subset \mathbb{R}^n \times \{0\}^{d-n}.
    \end{equation}
\end{theorem}

The submanifold property is comes from rank theorem stated generally below.

\begin{theorem}[Rank Theorem]
    Given manifolds $\mathcal{M}$ and $\mathcal{N}$ of dimension $n,d$, respectively, and a differentiable function $F$ of constant rank $r$ between them, then for every $p\in \mathcal{M}$, there are charts $(V_k, \psi_k)$ on $\mathcal{M}$ and $(U_k, \varphi_k)$ on $\mathcal{N}$ such that $p\in V_k$, $F(p)\in U_k$ such that $F(V_k) \subset U_k$ and 
    \begin{equation}
        \varphi_k \circ F \circ \psi_k^{-1} \, (x_1, \dots, x_r, x_{r+1},\dots, x_n) = (x_1, \dots, x_r, 0, \dots, 0).
    \end{equation}
\end{theorem}

In the above statement, if we consider $\mathcal{N}$ as $\mathbb{R}^d$ and $F$ as an inclusion map, then the rank $r$ is equal to $n$.
Also, we may ragard $( \mathcal{M} \cap U_k, \varphi_k |_{\mathcal{M} \cap U_k} )$ as a chart of $\mathcal{M}$ since it is compatible with $\{ (V_k,\psi_k) \}$ that provided by rank theorem.
This proves the submanifold property.

\subsection{Probabilistic background} \label{probability}

By a refined partition $\{ U'_{k'} \}_{k'=1}^{L'}$ of cover $\{ U_k \}_{k=1}^L$, we mean that it forms a partition of $\mathcal{M}$ itself and each $U'_{k'}$ is either a proper subset of some $U_k$ or intersection of $U_k$'s.
Here a proper subset of $U_k$ means a subset $U_k \setminus \cup \{ U_{k_0} : k_0 \neq k \}$.
Intuitively, $\{ U'_{k'} \}_{k'=1}^{L'}$ is obtained from $\mathcal{M}$ by cutting along all boundaries of $U_k$'s.
We define the quotient map $\pi : \mathcal{M} \to \{1,\dots,L'\}$ such that if $x$ is in $U_{k'}$ then $\pi(x) = k'$.
We let $\nu$ be a probability on $ \{1,\dots,L' \}$ which is defined by a push-forward measure of $p^\mathcal{M}$ by $\pi$.

\begin{definition}
In the above setting, there is a \textit{disintegration} of $p^\mathcal{M}$ given by
\begin{equation}
    p^\mathcal{M} 
    = \int_{ \{1,\dots,L'\} } p^{U'}_{k'} \; d\nu(k')
    = \sum_{k'=1}^{L'} \nu(k') p^{U'}_{k'},
\end{equation}
where $p^{U'}_{k'}$ is the probability on $U'_{k'}$ that can be considered as the probability on $\mathcal{M}$ with support in $U'_{k'}$.
\end{definition}

Since $p^\mathcal{M}$ is unknown, we cannot calculate its push-forward measure $\nu$.
Therefore, for each $k'=1,\dots,L'$, we estimate the value of $\nu(k')$ to the normalized number of sampled data points in $U'_{k'}$.
We note that $\nu(k') = p^\mathcal{M} (U'_{k'})$ by the definition of the push-forward measure.
And expectations of the numbers of sampled data points in $U'_{k'}$'s are proportional to $p^\mathcal{M} (U'_{k'})$, which is equal to $\nu(k')$.
Hence our estimation is suitable for sufficiently sampled datasets.

In order to estimate $p^\mathcal{M}$ on each cover $U_k$, we decompose $p^\mathcal{M}$ as
\begin{equation} \label{decomposition}
    p^\mathcal{M} = \sum_{k=1}^L c_k p^U_k,
\end{equation}
where $c_k$ is a scaling contant and $p^U_k$ is a probability on $\mathcal{M}$ with $p^U_k(U_k)=1$.
For the decomposition, we define $p^U_k$ as
\begin{equation}
    p^U_k = \frac{1}{c_k} \sum_{k':U_{k'} \subset U_k} \frac{1}{n(k')}  \nu(k') p^{U'}_{k'},
\end{equation}
where $U_{k'}$ consists of $n(k')$-intersection of $U_k$'s and 
\begin{equation}
    c_k = \sum_{k':U'_{k'} \subset U_k} \frac{1}{n(k')}  \nu(k')
\end{equation}
is the scaling constant that total measure of $p_\alpha$ is equal to 1.

In geometric terms, we can think of the decomposition of $p^\mathcal{M}$ to $p^U_k$'s in terms of "modified partition of unity".
We denote the probability density functions for $p^\mathcal{M}$ and $p^{U'}_{k'}$ by $\textbf{PDF}_\mathcal{M}$ and $\textbf{PDF}_{k'}$, respectively.
We define the probability density function $\textbf{PDF}_k$ supported on $U_k$ as
\begin{equation}
    \textbf{PDF}_k = \frac{1}{| \{ k'=1,\dots,L' : U'_{k'} \subset U_k \} |} \sum_{k': U'_{k'} \subset U_k} \textbf{PDF}_{k'}.
\end{equation}

\begin{definition}
Given a cover $\{ U_k \}_{k=1}^L$ on $\mathcal{M}$, a modified partition of unity is a family $\{ \rho_k : \mathcal{M} \to \mathbb{R} \}$ of functions such that for each $k=1,\dots,L$, the value of $\rho_k$ outside $U_k$ is equal to 0 and 
\begin{equation} \label{mpu}
    \int_{-\infty}^\infty \sum_{k=1}^L \rho_k (x) \; dx = 1.
\end{equation}
\end{definition}

The original definition of a partition of unity is that the left hand-side of (\ref{mpu}) does not have the integral.
For applying its concept to a probability density function, we modified the definition.
Then if we set a partition of unity as
\begin{equation}
    \rho_k = \sum_{k': U'_{k'} \subset U_k} \frac{1}{n(k')} \nu(k') \mathbf{1}_{U'_{k'}},
\end{equation}
where $\mathbb{1}$ is an indicator function, we have
\begin{equation}
    \sum_{k=1}^L \rho_k \textbf{PDF}_k
    = \sum_{k=1}^L \sum_{k': U'_{k'} \subset U_k} \frac{1}{n(k')} \nu(k') \textbf{PDF}_{k'}
    = \sum_{k': U'_{k'} \subset U_k} \nu(k') \textbf{PDF}_{k'}
    = \textbf{PDF}_\mathcal{M}
\end{equation}
which yields the same result with (\ref{decomposition}).

\subsection{Experiment details}\label{sec:exp_detail}

We specify the details of our experiments in this section. we use python of version $3.8$ for overall experiments.
For the essential packages in our algorithm, We use Kepler Mapper of version $2.0.1$, PyTorch of version $1.10.0$, and scikit-learn of version $1.0.1$.
We use Isomap and PCA from the scikit-learn package.

\subsubsection{Gaussian mixture model on a trefoil knot}
The trefoil knot without noise is parametrized by the following equations. For $t \in[0,2 \pi]$,
\begin{equation}
    \begin{cases}
        x=\sin (t)+3 \sin (2 t) \\
        y=\cos (t)-3 \cos (2 t) \\
        z=-\sin (3t).
    \end{cases}
\end{equation}
We add Gaussian noise with a mean of 0 and a standard deviation of $0.1$. The Gaussian mixture model consists of two Gaussian probability densities whose means are $0, \pi$, and standard deviations are all $\pi / 6$. We train the manifold learning flow and density estimation flow with $10^{4}$ data samples.

For the lens function of the Mapper algorithm, we use $1$-dimensional PCA values.
In the Mapper algorithm, Single-linkage clustering is used for clustering algorithm with distance threshold 1.
For the other hyperparameter for Mapper, n\_cubes is $2$, and perc\_overlap is $0.2$.

During every epoch, we use Adam optimizer with an initial rate of $0.0015$ decreasing the learning rate with cosine annealing with weight decay of $10^{-4}$. For gradient clipping, we set the norm $5$. We set the calculating step for expected points as $C_s=2$.
In the (Algorithm \ref{alg:all}), the number of epochs are $e_1, e_2, e_3, e_4, e_5 = 15, 30, 60, 60, 60$ respectively. The factors for weights are $\lambda_m, \lambda_p, \lambda_o, \lambda_d = 100, 0.01, 100, 0.1$, respectively. We set the batch sizes $b=256$.

\subsubsection{Gaussian mixture model on a torus}
The parametric equations for the torus are as follows.
For $t,s \in [0,2\pi]$,
\begin{equation}
    \begin{cases}
    x = ( \cos(t) + 3 ) \cos(s) \\
    y = ( \cos(t) + 3 ) \sin(s)  \\
    z = \sin(t).
    \end{cases}
\end{equation}
We add Gaussian noise with a mean of $0$ and a standard deviation of $0.1$.
The Gaussian mixture model consists of four Gaussian probability densities whose means are randomly chosen on $[-\pi,\pi]$ and standard deviations are all $\pi/3$.
The Gaussian mixture noise is added through the parameters $t, s$.
We train the models with $10^4$ data points from the noisy torus.

For the lens function of the Mapper algorithm, we use $1$-dimensional PCA values.
In Mapper algorithm, Single-linkage clustering is used for clustering algorithm with distance threshold 1.
For the other hyperparameter for Mapper, n\_cubes is $5$, and perc\_overlap is $0.45$.

During every epoch, we use Adam optimizer with an initial rate of $0.0015$ decreasing the learning rate with cosine annealing with weight decay of $10^{-4}$. For gradient clipping, we set the norm $5$. We set the calculating step for expected points as $C_s=2$.
In the (Algorithm \ref{alg:all}), the number of epochs are $e_1, e_2, e_3, e_4, e_5 = 60, 30, 60, 60, 60$ respectively. The factors for weights are $\lambda_m, \lambda_p, \lambda_o, \lambda_d = 100, 0.1, 25, 0.01$, respectively. We set the batch sizes $b=256$.

\subsubsection{StyleGAN2}

For both $S^2$ and $S^{511}$ experiments, we use $1$ dimensional PCA lens function in the Mapper algorithm.
For the other hyperparameter, the overlapping percentage is 0.33, and n\_cube is $2$.

During every epoch, we use Adam optimizer with an initial rate of $0.0001$ decreasing the learning rate with cosine annealing with weight decay of $10^{-4}$. For gradient clipping, we set the norm $1$. We set the calculating step for expected points as $C_s=5$. 
We noticed that the pretraining loss is not necessary.
In the \ref{alg:all}, the number of epochs are $e_1, e_2, e_3, e_4, e_5 = 0, 0, 40, 80, 60$ respectively. The factors for weights are $\lambda_m, \lambda_p, \lambda_o, \lambda_d = 100, 0, 100, 0.01$, respectively. We set the batch sizes $b=512$.

\newpage
\subsection{Algorithms}
$~$

\begin{algorithm}[h]
    \caption{Pretraining loss}
    \label{alg:pre}
    \begin{algorithmic}[1]
    \State {\bfseries Input:} Flow function $\varphi$.
    Data points $\{x_i\}_{i=1}^b$ on $\mathbb{R}^d$.
    Reference data points on $\mathbb{R}^n$, $R = \{r_i\}_{i=1}^b$.
    \State {\bfseries Output:} Pretraining loss $\mathcal{L}_{\text{pre}}$, 
    \For{$i = 1, \dots, b$}
        \State $v_i, \tilde{v}_i \leftarrow \varphi(x_i)$
    \EndFor
    \State $\mathcal{L}_{\text{pre}}$ = $\frac{1}{b}\sum_i \|v_i - r_i\|_2^2$
    \end{algorithmic}
\end{algorithm}

\begin{algorithm}[h]
    \caption{Reconstruction loss}
    \label{alg:recon}
    \begin{algorithmic}[1]
    \State {\bfseries Input:} Flow function $\varphi$.
    Data points $\{x_i\}_{i=1}^b$ on $\mathbb{R}^d$.
    \State {\bfseries Output:} Reconstruction loss $\mathcal{L}_{\text{recon}}$
    \For{$i = 1, \dots, b$}
        \State $v_i, \tilde{v}_i\leftarrow \varphi(x_i)$
        \State $x_i' \leftarrow \varphi^{-1}(v_i, 0)$
    \EndFor
    \State $\mathcal{L}_{\text{recon}}$ = $\frac{1}{b}\sum_i\|x_i - x_i'\|_2^2$
    \end{algorithmic}
\end{algorithm}

\begin{algorithm}
    \caption{Pairwise distance loss}
    \label{alg:pdl}
    \begin{algorithmic}[1]
    \State {\bfseries Input:} Flow function $\varphi$.
    Data points $\{x_i\}_{i=1}^b$ on $\mathbb{R}^d$.
    Reference pairwise distance matrix, $A = [a_{ij}]^{b\times b}$.
    \State {\bfseries Output:} Pairwise distance $\mathcal{L}_{\text{dist}}$
    \For{$i = 1, \dots, b$}
        \State $v_i, \tilde{v}_i \leftarrow \varphi(x_i)$
    \EndFor
    \State $T \leftarrow 0$
    \For{$i = 1, \dots b$}
        \For{$j = 1, \dots b$}
            \State $T \leftarrow T + (a_{ij} - \|v_i - v_j\|_2)^2$
        \EndFor
    \EndFor
    \State $\mathcal{L}_{\text{dist}}$ = $\frac{1}{b(b-1)}T$
    \end{algorithmic}
\end{algorithm}

\begin{algorithm}
    \caption{Compatibility loss}
    \label{alg:CompLoss}
    \begin{algorithmic}[1]
    \State {\bfseries Input:} Cover-wise coordinate maps, $\{\varphi_k\}_{k=1}^L$.
    Cover-wise data points $X_k = \{x_i^k\}_{i=1}^{b}$ on $\mathbb{R}^d$ and expected points $\{\hat{x}_i\}_{i=1}^{b}$.
    
    \State {\bfseries Output:} Compatibility loss $\mathcal{L}_{\text{comp}}$
    
    \State $T\leftarrow 0$
    \State $j\leftarrow 0$
    \For{$i = 1, \dots, b$}
        \If{$x_i^k$ is in overlapping region}
            \State $T \leftarrow T + \|x_i^k - \hat{x}_i\|_2^2$
            \State $j \leftarrow j + 1$
        \EndIf
    \EndFor
    \State $\mathcal{L}_{\text{comp}}$ = $\frac{1}{j}T$
    \end{algorithmic}
\end{algorithm}

\begin{algorithm}
    \caption{Density learning loss}
    \label{alg:dll}
    \begin{algorithmic}[1]
    \State {\bfseries Input:} Flow function $\gamma$. 
    Data points $\{v_i\}_{i=1}^b$.
    Probability density function on $\mathbb{R}^n$, $p^W$.
    \State {\bfseries Output:} $\mathcal{L}_{\text{density}}$, density loss
    \For{$i = 1, \dots, b$}
        \State $w_i \leftarrow \gamma(v_i)$
    \EndFor
    \State $\mathcal{L}_{\text{density}} \leftarrow -\frac{1}{b}\sum_i [\log p^W(w_i) - \log\det J_{\gamma^{-1}}(w_i)]$
    \end{algorithmic}
\end{algorithm}

\begin{algorithm}
    \caption{Calculate Expected Points}
    \label{alg:CEP}
    \begin{algorithmic}[1]
    \State {\bfseries Input:} All data points, $X = \{x_i\}_{k=1}^N$. Cover-wise data points $\{X_k\}_{k=1}^L$ .
    
    \State {\bfseries Output:} Expected points $\{\hat{x}_i\}_{i=1}^N$
    \For{$x_i$ in $X$}
        \State $\hat{x}_i \leftarrow 0$
        \State $j \leftarrow 0$
        \For{$k = 1, \dots,L$}
            \If{$x_i$ in $X_k$}
                \State $\hat{x}_i \leftarrow \hat{x}_i + x_i$
                \State $j \leftarrow j + 1$
            \EndIf
        \EndFor
        \State $\hat{x}_i \leftarrow \hat{x}_i / j$
    \EndFor
    \end{algorithmic}
\end{algorithm}

\begin{algorithm}
\caption{Manifold Learning Loss}
\label{alg:mfdl}
\begin{algorithmic}[1]
    \State{\bfseries Input:} Graph-based pairwise distance matrix of all data points, $D$. Flow function $\varphi$. Data points $\{x_i\}$. Factor weighting terms in the loss function, $\lambda_t$.
    \State{\bfseries Output:} Manifold Learning Loss $\mathcal{L}_{\text{mfd}}$
    \State $D_b\leftarrow $pairwise distance matrix corresponding the batch data $\{x_i\}$.
    \State $\mathcal{L_{\text{dist}} \leftarrow} \operatorname{Pairwise Distance Loss}(\varphi, \{x_i\}, D_b)$
    \State $\mathcal{L_{\text{recon}} \leftarrow} \operatorname{Reconstruction loss}(\varphi, \{x_i\})$
    \State $\mathcal{L_{\text{mfd}}}\leftarrow \lambda_t\mathcal{L}_{\text{dist}} + (1 - \lambda_t)\mathcal{L}_{\text{recon}}$
\end{algorithmic}
\end{algorithm}

\begin{algorithm}
\footnotesize
\caption{Training process}
\label{alg:all}
\begin{algorithmic}[1]
    \State {\bfseries Input:}  Cover-wise coordinate maps $\{\varphi_k\}_{k=1}^L$. Cover-wise density transforming map $\{\gamma_k\}_{k=1}^L$. Weights of $\{\varphi_k\}_{k=1}^L$, $\{\theta_k\}_{k=1}^L$. Weights of $\{\gamma_k\}_{k=1}^L$, $\{\omega_k\}_{k=1}^L$. The learning rate, $\alpha$. The batch size, $b$. All point cloud, $X$. Cover-wise point clouds, $\{X_k\}_{k = 1}^L$. Cover-wise Isomap results of $\{X_k\}_{k=1}^L$, $\{R_k\}_{k=1}^L$. Cover-wise graph-based pairwise distance matrix of $\{X_k\}_{k=1}^L$, $\{D_k^{N_k\times N_k}\}_{k=1}^L$. Pair-wise probability density function on space $W$, $\{p_k^W\}_{k=1}^L$. Factors weighting terms in the loss functions, $\lambda_m, \lambda_p, \lambda_o$ and $\lambda_d$. The number of epochs for each step, $e_1, e_2, e_3, e_4$ and $e_5$. The number of steps for calculating expected points, $C_s$.
    \For{$k = 1,\dots L$}
        \For{$j=1, \dots, e_1$}
            \For{mini-batch data $\{x_i\}_{i=1}^b, \{r_i\}_{i=1}^b$ from $X_k$ and $R_k$ respectively}
                \State $\mathcal{L_{\text{pre}} \leftarrow} \operatorname{Pretraining loss}(\varphi_k, \{x_i\}, \{r_i\})$
                \State $\theta_k\leftarrow \theta_k - \alpha \lambda_m \nabla_{\theta_k}\mathcal{L}_{\text{pre}}$
            \EndFor
        \EndFor
        
        \For{$j= 1, \dots, e_1$}
            \For{mini-batch data $\{x_i\}_{i=1}^b$ from $X_k$}
                \State $L_{\text{density}}\leftarrow\operatorname{Density Learning Loss}(\gamma_k, \{\varphi_k(x_i)\}, p_k^W )$
                \State $\omega_k\leftarrow \omega_k - \alpha \lambda_d \nabla_{\omega_k}\mathcal{L}_{\text{density}}$
            \EndFor
        \EndFor
        \For{$j=1, \dots, e_2 + e_3$}
            \If{$j \leq e_2$}
                \State $\lambda_t \leftarrow \frac{\lambda_p - 1}{e_2 - 1} j + 1 - \frac{\lambda_p-1}{e_2 - 1}$
            \EndIf
            \For{mini-batch data $\{x_i\}_{i=1}^b$ from $X_k$}
                \State $\mathcal{L_{\text{mfd}}}\leftarrow \operatorname{Manifold Learning Loss}(D, \varphi_k, \{x_i\}, \lambda_t)$
                \State $\theta_k\leftarrow \theta_k - \alpha \lambda_m \nabla_{\theta_k}\mathcal{L_{\text{mfd}}}$
                
                \State $L_{\text{density}}\leftarrow\operatorname{Density Learning Loss}(\gamma_k, \{\varphi_k(x_i)\}, p_k^W )$
                \State $\omega_k\leftarrow \omega_k - \alpha \lambda_d \nabla_{\omega_k}\mathcal{L}_{\text{density}}$
            \EndFor
        \EndFor
    \EndFor
    
    \For{$j = 1, \dots, e_4$}
        \If{$j\, \% \,C_s$ is $1$}
            \State $\{\hat{x}_i\} \leftarrow\operatorname{Calculate Expected Points}(X, \{X_k\}_{k=1}^L)$
        \EndIf
        \For{$k=1, \dots L$}
            \For{mini-batch data $\{x_i\}_{i=1}^b$ from $X_k$}
                \State $\mathcal{L_{\text{mfd}}}\leftarrow \operatorname{Manifold Learning Loss}(D, \varphi_k, \{x_i\}, \lambda_p)$
                \State $\mathcal{L_{\text{comp}} \leftarrow} \operatorname{Compatibility Loss}(\{\varphi_k\}, \{X_k\}, \{\hat{x}_i\})$
                \State $\mathcal{L}\leftarrow \mathcal{L}_{\text{mfd}} + \frac{1}{e_4}j\lambda_o\mathcal{L}_{\text{comp}}$
                
                \State $\theta_k\leftarrow \theta_k - \alpha \lambda_m \nabla_{\theta_k}\mathcal{L}$
                
                \State $L_{\text{density}}\leftarrow\operatorname{Density Learning Loss}(\gamma_k, \{\varphi_k(x_i)\}, p_k^W )$
                \State $\omega_k\leftarrow \omega_k - \alpha \lambda_d \nabla_{\omega_k}\mathcal{L}_{\text{density}}$
            \EndFor

        \EndFor
    \EndFor
    \For{$k = 1,\dots, L$}
        \For{$j = 1, \dots, e_5$}
            \For{mini-batch data $\{x_i\}_{i=1}^b$ from $X_k$}
                \State $L_{\text{density}}\leftarrow\operatorname{Density Learning Loss}(\gamma_k, \{\varphi_k(x_i)\}, p_k^W )$
                \State $\omega_k\leftarrow \omega_k - \alpha \lambda_d \nabla_{\omega_k}\mathcal{L}_{\text{density}}$
            \EndFor
        \EndFor
    \EndFor
\end{algorithmic}
\end{algorithm}

\end{document}